\documentclass[10pt,twocolumn,letterpaper]{article}

\usepackage[pagenumbers]{cvpr} 

\usepackage{graphicx}
\usepackage{amsmath}
\usepackage{tabularx}
\usepackage{amssymb}
\usepackage{booktabs}
\DeclareMathOperator*{\argmax}{arg\,max}
\DeclareMathOperator*{\argmin}{arg\,min}
\usepackage{algorithmicx}
\usepackage{algorithm}
\usepackage{algpseudocode}
\usepackage{sidecap, caption}

\usepackage[pagebackref,breaklinks,colorlinks]{hyperref}

\usepackage[capitalize]{cleveref}
\crefname{section}{Sec.}{Secs.}
\Crefname{section}{Section}{Sections}
\Crefname{table}{Table}{Tables}
\crefname{table}{Tab.}{Tabs.}

\def\cvprPaperID{1247} 
\def\confName{CVPR}

\begin{document}

\title{Adversarially Robust Video Perception by Seeing Motion}



\newcommand*\samethanks[1][\value{footnote}]{\footnotemark[#1]}
\author{
Lingyu Zhang\thanks{Equal Contribution. Order by coin flip.}
\hspace{1em}
Chengzhi Mao\samethanks
\hspace{1em} Junfeng Yang\hspace{1em}Carl Vondrick
\\
Columbia University\\
{\tt\small  lz2814@columbia.edu, \{mcz,junfeng,vondrick\}@cs.columbia.edu}}

\maketitle




\begin{abstract}
Despite their excellent performance, state-of-the-art computer vision models often fail when they encounter adversarial examples. Video perception models tend to be more fragile under attacks, because the adversary has more places to manipulate in high-dimensional data. In this paper, we find one reason for video models' vulnerability is that they fail to perceive the correct motion under adversarial perturbations. Inspired by the extensive evidence that motion is a key factor for the human visual system, we propose to correct what the model sees by restoring the perceived motion information. Since motion information is an intrinsic structure of the video data, recovering motion signals can be done at inference time without any human annotation, which allows the model to adapt to unforeseen, worst-case inputs. Visualizations and empirical experiments on UCF-101 and HMDB-51 datasets show that restoring motion information in deep vision models improves adversarial robustness. Even under adaptive attacks where the adversary knows our defense, our algorithm is still effective. Our work provides new insight into robust video perception algorithms by using intrinsic structures from the data. Our webpage is available at \href{https://motion4robust.cs.columbia.edu/}{motion4robust.cs.columbia.edu}.


\end{abstract}

\section{Introduction}
\label{sec:intro}

Deep models have consistently achieved high performance over a large number of computer vision tasks~\cite{alexnet, vgg, maskrcnn, 3dcnn, convspatio-temp}. However, it has been shown that they are susceptible to adversarial examples\cite{Carlini2017TowardsET, Goodfellow2015ExplainingAH, Szegedy2014IntriguingPO}, where additive perturbations are optimized to fool the model. The existence of adversarial examples induces real threats to real-world scenarios where safety and robustness are crucial~\cite{advphysical, attackcars, attackface}. To address this problem, a large body of work has studied adversarially robust models for image recognition~\cite{madry2018towards, Goodfellow2015ExplainingAH, mao19metric, zhang2019theoretically, cohen2019certified, Mao_2021_ICCV, nie2022diffusion, wu2021attacking}. While robustness on static images achieved signifiantly progress, robust models on the more realistic video data with extra time dimension are still under-explored~\cite{Xiao2019AdvITAF}.



Learning robust video perception models against adversarial attacks is challenging because models suffer when the input data has a high number of dimensions~\cite{SimonGabriel2019FirstOrderAV}. Adversarial training~\cite{Goodfellow2015ExplainingAH, madry2018towards} has been studied to increase video models' robustness \cite{kinfu2022advtrainvid}, yet they achieved less robustness than training on images. In addition, since they are training-time defense, they need to anticipate the type of adversarial attacks at inference time in the first place and then adversarially train on them, which causes them to be vulnerable to unforeseen attacks \cite{zhang2019limitations}. In addition, adversarially train a robust video model is orders of times more expensive than training a standard video model, which limits the deployment of training-based methods.







\begin{figure*}[t]
 \centering
  \includegraphics[width=0.9\textwidth]{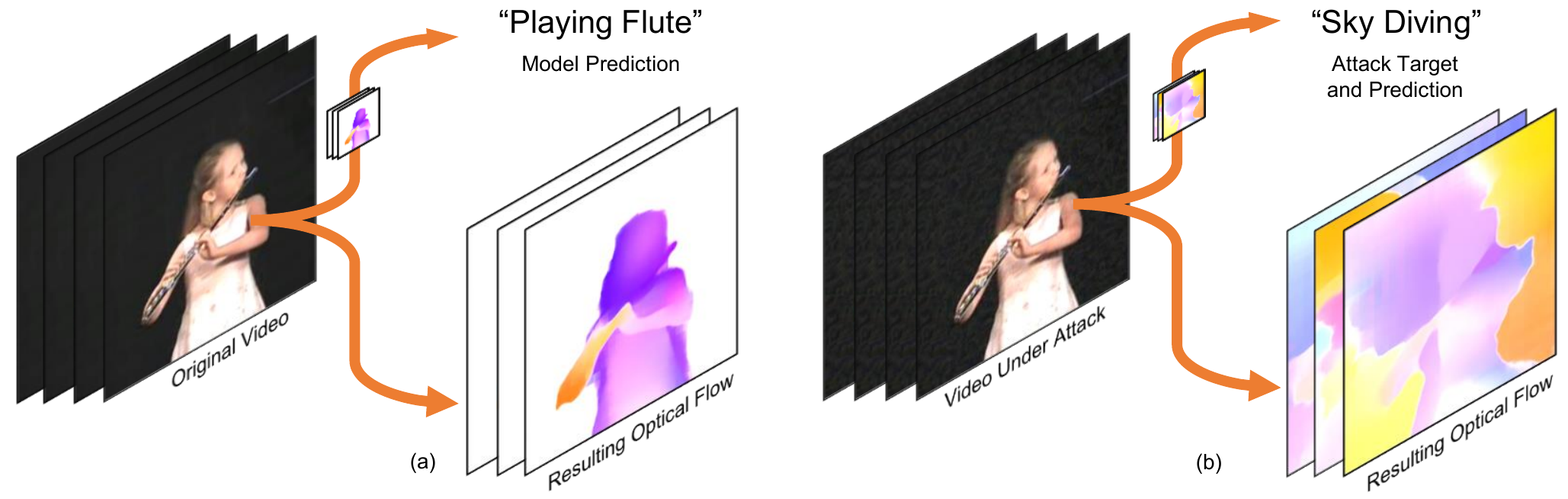}
  \caption{Adversarial attacks for video perception models often end-to-end optimize the attack noise to fool the classification results. Despite the attack never directly optimizing to change the optical flow, we find they also collaterally corrupt the optical flow, where the attack corrupts the flow from (a) into (b). Our work aims to exploit the inconsistencies between the optical flow and the video in order to defend against attacks.}
  \label{fig:teaser}
\end{figure*}

 If you want your friend to find you far away, you wave your arms to stand out from the background. Human vision often sees moving things well. The ``law of common fate'' from Gestalt psychology~\cite{koffka1922perception} also shows that things that move together often belong to the same group. In computer vision, motion has also been shown to be robust in terms of cross-domain generalization as they are appearance-invariant \cite{plizzari2022egomotion, sevilla2018integration}. This motivates us to capitalize on motion clues to build robust video perception models. 
 
 One common practice is to use optical flow as the input representation for the video perception models~\cite{ng2018actionflownet, sevilla2018integration, Inkawhich2018AdversarialForOptical}. We start our investigation by analyzing the quality of motion---such as optical flow---used in a deep model. While the quality of estimated optical flow is high in clean videos, we find optical flow is corrupted by the adversarial examples, even though the adversarial noise is solely optimized to fool the final category prediction (see Figure~\ref{fig:teaser}). This suggests that the video perception model fails under attacks partially because they perceive inaccurate motion. We hypothesize that if we can enable the model to perceive the right motion, we can obtain a robust video classifier.

\begin{figure*}[t]
 \centering
  \includegraphics[width=0.7\textwidth]{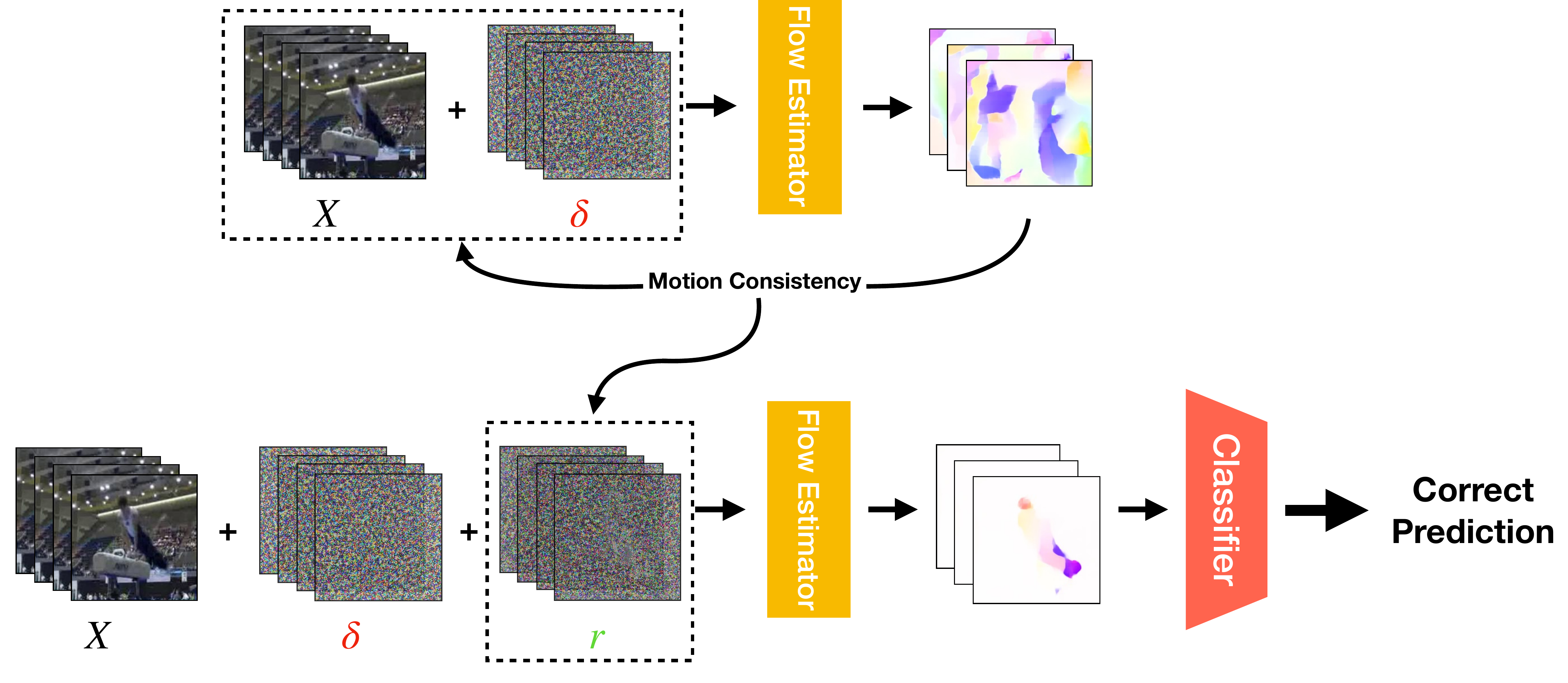}
  \caption{{\bf Our method.} Adversarial perturbations ($\delta$) on RGB inputs result in inaccurately estimated motion, indicated by the distorted optical flow. Distorted optical flow provides a self-supervised consistency loss measured by warping error. Optimizing for this consistency allows us to obtain a reverse perturbation $r$ that restores the temporal consistency and corrects misclassified results.}
  \label{fig:method}
\end{figure*}

While one can adversarially \emph{train} the optical flow model against attacks, the model's ability to perceive the correct motion is not guaranteed at \emph{test} time, because the optical flow may still be subverted under stronger, unforeseen attacks. Instead, we propose to make the model perceive the correct motion at \emph{inference} time. Our key insight is that after warping one video frame with the observed motion, it should be able to accurately construct the next frame. Since adversarial attacks corrupt the motion, it creates inconsistency between the warped frame and the true frame, providing us with a signal to adjust the video input, as illustrated in \cref{fig:method}. We then adapt the input so that this motion consistency is respected, which corrects the optical flow estimation. Our objective does not require any annotation and can be performed dynamically at inference time.


Visualizations and empirical results show that enforcing motion improves models' robustness under four different adversarial attacks on two datasets, UCF-101, and HMDB-51, by up to 72 points. Since our method is at inference time, the user can choose whether to perform this defense based on need.
Even if the attacker knows our defense and adapts their attack to break ours, our method still remains robust, improving robustness by over 7 points. We will release our model, data, and code.

\section{Related Work}
\label{sec:related}


{\bf Motion Estimation}. Optical flow is the standard way to capture motion. Classical variational methods\cite{horn1981determining, lucas1981iterative, memin1998denseOF, brox2004highacc, TV-L1} for estimating optical flow are based on the minimization of an energy function, which typically consists of a photometric consistency term and a smoothness term.
Recently, convolutional neural network-based methods demonstrate better optical flow estimation\cite{dosovitskiy2015flownet, flownet2, sun2018pwc, teed2020raft}. These approaches usually directly regress the ground truth optical flow calculated from synthetic data in a supervised manner.
More closely related to our work, deep unsupervised optical flow estimation has also been an interesting line of research \cite{jonschkowski2020matters}. They optimize for a similar objective function to that of variational methods, while adopting architectures from supervised neural optical flow estimation as backbones.

{\bf Adversarial Attacks and Defenses on Videos.} Since the discovery of neural networks' vulnerability to adversarial examples, various types of attacks and defenses on static images have been extensively studied. Not until recently did a few studies start investigating adversarial videos. \cite{Wei2019SparseAP} proposed a perturbation regularized by the $l_{2,1}$ norm that induces sparsity on the adversarial noise. \cite{wei2020blackbox} designed a black-box attack based on heuristically found important frames. Similarly, the attack proposed by \cite{Hwang2021JustOM} also relies on the pre-discovery of critical frames in videos, but they take it further to only perturbing one frame in a white-box setting. \cite{Pony2021OvertheAirAF} on the other hand constructed unconstrained attacks by adding a constant flickering color shift to each frame, obtaining more humanly-imperceptible attacks.

Defense for videos is more challenging and less explored in the literature. \cite{Xiao2019AdvITAF} made the first effort to identify adversarial frames using temporal consistency, without defending them. \cite{Jia2019IdentifyingAR} proposed a detect-then-defend strategy that classifies the type of attack first and then applies different types of defense accordingly. However, they did not test their defense against any adaptive attacks. \cite{lo2021unforeseenvideos} have a similar adversarial detector component, and proposed to apply multiple independent batch normalization to improve robustness under different types of distribution shifts caused by adversarial perturbations. Their approach requires adversarial training and is only evaluated against weak adversarial attacks. 

{\bf Inference time adversarial defense.} Adversarial training\cite{Goodfellow2015ExplainingAH, kurakin2017atscale, madry2018towards} is a standard technique for obtaining robust models. By training on worst-case examples, the models learn to be invariant to adversarial perturbations. However, the robustness decreases when they encounter attacks they were not trained on. Another problem is that adversarial training often leads to significant degradation in clean performance~\cite{tsipras2018robustness}. Adversarial training for videos is even more difficult with the growing amount of data and increased dimensionality, because it is easier to construct attacks that not showed in training time in a higher dimensional space.

Recently, inference-time defense has been shown to be effective in improving robustness\cite{kang2021stable, wu2021attacking, Mao_2021_ICCV, liu2022landscape, nie2022diffusion}, shifting the burden of robustness from training to testing. The model performs robust latent inference by optimizing for specific energy functions~\cite{mao2022robust, Mao_2021_ICCV, wang2019bidirectional}. This allows the model to adapt to the unforeseen characteristics of attacks and corruptions at inference time. Since there is no need to retrain the models, inference-time defenses work well with the vast number of available pre-trained models. 


\section{Motion for Robust Perception}

In this section, we first provide background for motion-based action recognition models that uses optical flow as input representations. We then show that adversarial attacks collaterally damage the motion structure in videos. We present our defense method that reverses adversarial attacks by restoring the motion consistency at inference time.

\subsection{Background: Optical Flow for Perception}
\label{sec:flow}

In video understanding, models that use the right motion signals often obtain better robustness \cite{plizzari2022egomotion, sevilla2018integration}. Optical flow is a popular method to capture motion. Optical flow explicitly encodes the movement in videos by calculating a field of displacement vectors of corresponding pixels between each frame. Since optical flow can be calculated for every pixel in every frame, it provides rich signals for video understanding. 

State-of-the-art action recognition models take in optical flow fields as input \cite{simonyan2014twostream, feichtenhofer2016convolutional, sevilla2018integration, carreira2017i3d}, where optical flow are precomputed with the traditional variational methods, such as the TV-L1\cite{TV-L1} algorithm. However, these approaches require many iterations to achieve accurate flows and are slow at actual test time. In this work, we chose the popular neural network-based optical flow estimators, since they obtain state-of-the-art performance on both accuracy and speed.


 We define our model as the following. Let $X=\{x_1, ... x_{T}\}$ be the input video clip with $T$ frames. For every pair of adjacent frames $x_i, x_{i+1}$, an optical flow field $f_{i}$ is estimated using a pre-trained neural flow estimator $G$: $f_{i} = G(x_{i}, x_{i+1})$. Let $F=\{f_1, ..., f_{T-1}\}$ be the stack of $T-1$ flow fields estimated from $X$. For simplicity, denote $F=G(X_{<T}, X_{>1})$, where $X_{<T}$ and $X_{>1}$ are the stack of all the first frames and second frames in every pair of adjacent frames. If not further specified, let $G(X) = G(X_{<T}, X_{>1})$ denote the forward direction flow estimator. $F$ is then passed through a classifier $H$ with a softmax layer at the end, outputting a categorical distribution over the classes $H(G(X))$. 

Following the best practice, we incorporate neural flow models for video classification~\cite{ng2018actionflownet, sevilla2018integration, Inkawhich2018AdversarialForOptical}.
We choose our architecture to be the combination of RAFT~\cite{teed2020raft} and I3D~\cite{carreira2017i3d} model, which are the state-of-the-art neural optical flow estimator and video classifier backbone, respectively.





\subsection{Test-Time Motion Consistency}
\label{sec:3.2}


We find that when an adversarial perturbation is applied to the input RGB clip $X$, the estimated flow fields $F$ would also be corrupted (as shown in Figure~\ref{fig:teaser}). Note that the attack vector is only optimized to fool the target classifier, such as action recognition.
However, we find the attack also collaterally damage the motion signals in the model. 

 Since motion clue is a key factor for visual perception, wrong motion signals will lead to inaccurate prediction. Our key insight is that motion is an inherent video signal, which we can recover even without annotation. This allows us to restore the motion signal at inference time, because obtaining motion does not require any annotation.

\begin{figure*}[h]
 \centering
  \includegraphics[width=\textwidth]{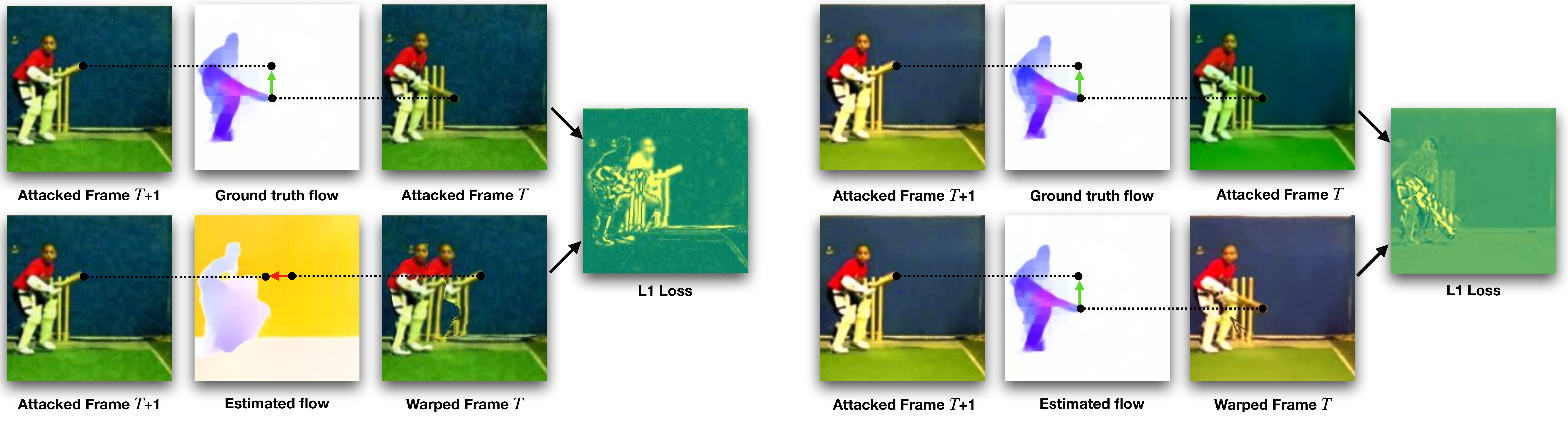}
  \caption{{\bf Our consistency objective can improve robustness in twofold:} (1) it enforces accurate optical flow (left) and (2) it suppresses pixel intensity distortion (right). If the estimated flow is highly corrupted by the attack, as shown on the left, the pixels in the $T+1$ frame will be warped to the wrong location in frame $T$. The $L_1$ distance between the warped and actual frame $T$ will be high, providing a signal to fix the optical flow. If the attacker shifts the pixel values too far from the original (\ie under flickering attacks\cite{Pony2021OvertheAirAF}), even if flow fields warp pixels to the correct locations, the $L_1$ distance will still be high. This provides signals to fix discontinuity in the data due to attack.}
  \vspace{-5mm}
  \label{fig:twofold}
\end{figure*}


There are a number of work estimation optical flow without annotation, where\cite{jonschkowski2020matters} conducted a comprehensive study for them. In this work, we consider two of the most prominent ones: photometric consistency and smoothness constraints.

\noindent{\bf Photometric consistency} (also known as warping error) is based on the constant brightness assumption \cite{horn1981determining}, \ie the pixel intensity of a scene point does not change when moving through frames. Formally, if a pixel in frame $t$ at location $(x,y)$ has intensity $I_t(x,y)$ and the displacement vector of this point given by the optical flow is $(u,v)$, then the pixel in the $t+1$ that is being pointed at should ideally have the same intensity. We define the photometric consistency loss of a video clip by summing over the consistency loss of every adjacent pair of frames:
\begin{equation} \label{eq1}
    \mathcal{L_\text{photo}} = \frac{1}{n} \sum_t \sum_{x,y}|I_{t+1}(x+u, y+v) - I_t(x, y)|_p
\end{equation}

Where $n$ is the number of pixels in one frame. This can be implemented in a differentiable manner by applying the warping operation $w$ on the second frames. If the optical flow is perfectly accurate, the warped frame should be consistent with the first frame. We can thus compute a similarity metric (not limited to $L_p$ distances) between the warped second frame and the first frame as a measurement of flow consistency: 

\begin{equation} \label{eq2}
    \mathcal{L_\text{photo}} = \sum_t \mathcal{L_{\text{sim}}} (w(I_{t+1}), I_t)
\end{equation}


\noindent{\bf Smoothness constraint} regularizes the estimated flow to have lower spatial frequency and be consistent with neighboring pixels. In this work, we follow \cite{jonschkowski2020matters} and define it by the edge-aware second derivative of the estimated flow:
 
\begin{equation} \label{eq3}
\begin{aligned}
    \mathcal{L_\text{smooth}} = \frac{1}{n}\sum \exp{(-\frac{\lambda}{3}\sum_c |\frac{\partial{I_c}}{\partial{x}}}|) |\frac{\partial ^2{V}}{\partial{x^2}}|\\
    + \exp{(-\frac{\lambda}{3}\sum_c |\frac{\partial{I_c}}{\partial{y}}}|) |\frac{\partial ^2{V}}{\partial{y^2}}|
\end{aligned}
\end{equation}

Where $\lambda$ controls the edge weighting, $I_c$ is the pixel intensities of color channel $c$ and $V$ is the estimated optical flow. Given an optical flow estimator $G$ and a similarity metric $\mathcal{L_\text{sim}}$, we hereby define the Motion Consistency ({\bf MC}) measurement:

\begin{equation} \label{eq4}
\begin{aligned}
\mathcal{L_{\text{MC}}}(X, G, \mathcal{L_\text{sim}}) =
&\mathcal{L_{\text{sim}}}(X_{<T}, W\cdot[X_{>1}, G(X)]) \\
+ &\lambda_{\text{smooth}} \cdot \mathcal{L_{\text{smooth}}}(X, G(X))
\end{aligned}
\end{equation}

Where W is the warping operator for a set of frames. $X_{<T}$ and $X_{>1}$can be seen as the first frames and second frames in all the pairs of adjacent frames. Now that we have defined the Motion Consistency objective, we can formulate a defense algorithm. Typically, an attacker optimizes for the following objective:

\begin{equation} \label{eq5}
\begin{aligned}
\delta^* &= \argmax_{\delta} \mathcal{L_{\text{CE}}}(y, H(G(X+\delta)))\\
& s.t. |\delta|_{p} \leq \epsilon_a
\end{aligned}
\end{equation}

In which $\mathcal{L_{\text{CE}}}$ is the cross-entropy loss, $y$ is the true label and $\epsilon_a$ is the attack budget. Our inference time defense optimizes for a reverse perturbation that minimizes the self-supervised {\bf MC} loss when added to the potentially attacked video:

\begin{equation} \label{eq6}
\begin{aligned}
r^* &= \argmin_{r} \mathcal{L_{\text{MC}}}(X+r) \\
& s.t. |r|_{p} \leq \epsilon_r
\end{aligned}
\end{equation}

The reverse perturbation bound $\epsilon_r$ is necessary, as without it could result in trivial solutions (\eg constant color frames would have zero {\bf MC} loss). We optimize the objective using projected gradient descent. Our algorithm is described in \cref{algorithm: reverse}.

\begin{algorithm}[t] 
\caption{Robust Inference that Respects Motion}
\label{algorithm: reverse}
\begin{algorithmic}[1]
\State {\bfseries Input:} Potentially attacked video clip $X$ with $T$ frames, step size $\eta$, number of iterations $K$, flow estimator $G$, video classifier $H$, reverse attack bound $\epsilon_r$, similarity function $\mathcal{L_\text{sim}}$, smooth loss function $\mathcal{L_\text{smooth}}$ and smooth loss weight $\lambda_\text{smooth}$.
\State {\bfseries Output:}  Prediction $\hat{y}$
\State {\bfseries Inference}: \\
$X'\leftarrow X$
\For{$k=1,...,K$}
    \State $F \leftarrow G(X')$ \algorithmiccomment{compute optical flow}
    \State $\hat{X'}_{[1...T-1]} \leftarrow W(F, X'_{[2..T]})$ \algorithmiccomment{backward warping}
    \State $\mathcal{L}_\textbf{MC} = \mathcal{L}_\text{sim} (\hat{X'}_{[1...T-1]}, {X'}_{[1...T-1]})$\\
     $\qquad \qquad +\lambda_\text{smooth}\mathcal{L_\text{smooth}}(X', F) $ \\ \algorithmiccomment{compute motion loss}
    \State $X'\leftarrow X' - \eta \cdot \text{sign}(\nabla_{X'} \mathcal{L}_\text{MC})$
    \State $X'\leftarrow \Pi_{(X,\epsilon_r)} X' $ \algorithmiccomment{project back into the bound}
\EndFor
\State Predict the final output by $\hat{y}=H(G(x'))$
\end{algorithmic}
\end{algorithm}

\subsection{Multiple Motion Constraints For Robustness}

It is shown that multitask and multiple constraints~\cite{mao2020multitask, lawhon2022using, mao2022robust} are pivotal for adversarial robustness. In \cref{sec:3.2} we discussed photometric consistency and smoothness as constraints for optical flow restoration. In natural videos, there are rich intrinsic constraints that the model should satisfy.  In previous sections, we used forward flow computation and backward warping:

\begin{equation}
 \mathcal{L_\text{forward}} = \mathcal{L_{\text{sim}}}(X_{<T}, W\cdot[X_{>1}, G(X_{<T}, X_{>1})])
\end{equation}

We can also compute backward flow with forward warping as an additional constraint. 

\begin{equation}
 \mathcal{L_\text{backward}} = \mathcal{L_{\text{sim}}}(X_{>1}, W\cdot[X_{>T}, G(X_{>1}, X_{<T})])
\end{equation}

In practice, this can be implemented by simply generating flows after reversing the order of the frames. Additionally, motion consistency should continue to hold for longer ranges across time. We can generate flows between frames that are not adjacent, requiring flows to be accurate for larger movements. Here we simply split each clip into $X_{<\frac{T}{2}}$ and $X_{\geq \frac{T}{2}}$, and compute flows between $X_i$ and $X_{i + \frac{T}{2}}$ for $i=0,...,\frac{T}{2}-1$. Therefore, the long-range photometric consistency is defined by:

\begin{equation}
 \mathcal{L_\text{long}} = \mathcal{L_{\text{sim}}}(X_{<\frac{2}{T}}, W\cdot[X_{\geq \frac{T}{2}}, G(X_{<\frac{2}{T}}, X_{\geq \frac{T}{2}})])
\end{equation}

Now we define the Multiple Motion Consistency objective:

\begin{equation}
    \mathcal{L_\text{multiMC}} = \mathcal{L_\text{forward}} + \mathcal{L_\text{backward}} + \mathcal{L_\text{long}} + \lambda_\text{smooth} \cdot \mathcal{L_\text{smooth}}
\end{equation}

In principle, one could continue introducing more constraints, such as photometric consistency between frames for every stride length. However, the computational expense grows linearly with the number of constraints, so in this work, we only consider the constraints above. We show in experiments that this is sufficient to yield improved robustness against defense-aware attacks.

\begin{figure*}[h]
 \centering
  \includegraphics[width=0.9\textwidth]{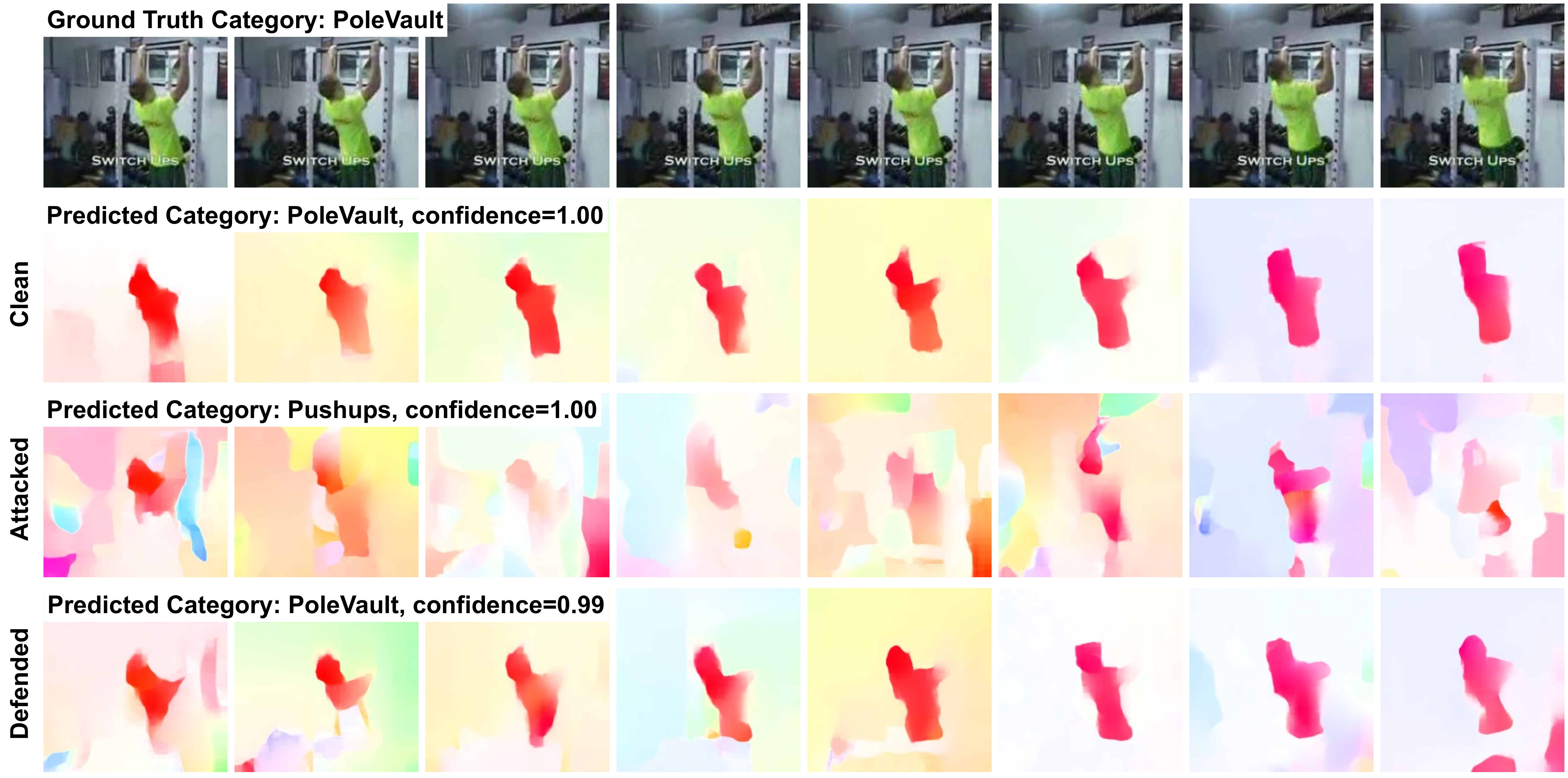}
  \caption{{\bf Visualization of attack and defense.} The first row is the frames of a video of the class "PoleVault". Row 2, 3 and 4 are the optical flow fields estimated on clean, attacked, and purified inputs, respectively. The adversarial attack on RGB inputs resulted in the wrong classification prediction "Pushups" as well as corrupted flow estimations. Using our defense method, we restore the accurate flow fields as well as the categorical prediction of the model.}
  \label{fig:dense}
  \vspace{-5mm}
\end{figure*}

\subsection{Defense-aware Attacks}

In a more realistic setting, the attacker could have knowledge of the defense being deployed, and adjust their strategy accordingly. In this section, we describe three types of defense-aware attacks targeting at our model.

\noindent{\bf Adaptive Attack I.}
One way a defense-aware attacker could make their attack more effective is to respect the defender's loss in their attack objective. As described in \cite{Mao_2021_ICCV}, one could construct an attack that maximizes the cross-entropy loss while minimizing the defender's loss at the same time by introducing a Lagrangian multiplier.

\begin{equation}
    \delta_{\bf{AA1}} = \argmax_{\delta} \mathcal{L_{\text{CE}}} - \lambda \mathcal{L_{\text{MC}}}
\end{equation}

\noindent By incorporating the defender's loss as a regularization, the attacker finds examples that are more likely to circumvent the defense.

\noindent{\bf Adaptive Attack II.}
The attack described above is a general way to counter defenses. We design another attack that specifically aims at bypassing our defense. Recall the observation in \cref{sec:flow}. The assumption that our defense relies on is that adversarial perturbations in the RGB space result in corrupted motion estimation. If an attacker was aware of our defense, they could apply attacks has minimal effects on the estimated motion by adding the change of flow as a regularization term.
\begin{equation}
    \delta_{\bf{AA2}} = \argmax_{\delta} \mathcal{L_{\text{CE}}} - \lambda |G(x+\delta) - G(x)|_p
\end{equation}

\noindent By doing so, the attack becomes more "stealthy" and provides a weaker signal for our defense.

\noindent{\bf Adaptive Attack III.}
A common issue for inference-time defenses is the reliance on obfuscated gradients. As pointed out in \cite{croce2022evaluating}, the iterative process for purification potentially leads to vanishing gradients, which provides a false sense of security as the attacker can deploy gradient approximation attacks, known as BDPA\cite{athalye2018obfuscated}, to circumvent this. In our experiments, we follow \cite{croce2022evaluating}, applying full forward pass for the purification, and replacing its gradients with the identity function during backward pass.

\section{Experiments}

We demonstrate the effectiveness of our defense against various types of attacks. We first show our model's capability of recovering performance under common strong dense attacks. A few attacks designed specifically for videos have been proposed in recent years. We analyze the performance of our defense against two state-of-the-art video attacks as well. Finally, we report results under various defense-aware attacks.

\subsection{Experimental settings}

\begin{table*}
\centering
\small
    \begin{tabular}{c|c|c|c|c|c|c}
         \toprule
         {} & \multicolumn{3}{c|}{UCF-101}  & \multicolumn{3}{c}{HMDB-51} \\
          Model   & Standard & Random & Defended  &  Standard & Random & Defended   \\
         \midrule 
         Clean & 86.6 & - & 84.0 & 67.0 & - & 66.9 \\
         Random & 79.9 & - & 82.2 & 62.7 & - & 63.9\\
         PGD ($\epsilon=4$) & 14.7 & 15.3 & \bf{78.5} & 8.8 & 8.8 & \bf{59.5} \\
         PGD ($\epsilon=8$) & 10.5 & 10.9 & \bf{73.9} & 7.9 & 7.7 & \bf{53.9}  \\
         PGD ($\epsilon=16$) & 6.2 & 6.7 & \bf{54.5} & 5.2 & 6.6 & \bf{40.8} \\
         AutoAttack ($\epsilon=8$) & 2.4 & 6.9 & \bf{75.3} & 1.3 & 5.6 & \bf{57.9} \\
         \bottomrule
    \end{tabular}
\caption{\small{Adversarial robust accuracies against dense attacks on UCF-101 and HMDB-51. By restoring the intrinsic motion information, our defense consistently improves performance under various types and strengths of dense attacks. Our method also preserves clean accuracies. With a cost of fewer than 3 points drop in clean accuracy, we obtain a robustness gain by up to 72 points.
}}
\vspace{-1mm}
\label{pgd}
\end{table*}

\noindent {\bf Datasets}. We evaluate the effectiveness of our method on two standard action recognition datasets UCF-101\cite{soomro2012ucf101}, and HMDB-51\cite{kuehne2011hmdb51}. Experiments are conducted on a subset of 1000 clips of the test sets.

\noindent {\bf Model under attack}. We chose the state-of-the-art RAFT\cite{teed2020raft} as our flow estimator front-end and the flow stream of the state-of-the-art I3D\cite{carreira2017i3d} as the classifier backbone. As reported by the original authors, RAFT has strong generalization across datasets. We directly use the RAFT model pre-trained on the Sintel dataset \cite{butler2012sintel}, as it was already capable of producing high-quality flows on UCF-101. We then fine-tune I3D models pre-trained on Kinetics \cite{carreira2017i3d} to our datasets.

\noindent {\bf Evaluation Details}. We resize all videos to $300\times400$ pixels and center crop $224\times224$ to eliminate black surrounding frames. We randomly sample 64-frame clips from videos and loop the ones that are too short. Note that we evaluate on videos with the largest input dimensions compared to prior works, which is more difficult to defend.

\noindent  Note that the recurrent update module in RAFT has the potential to create obfuscated gradients, which has been shown to lead to a false sense of security \cite{athalye2018obfuscated}. We circumvent this by approximating gradients during attack and defense. Specifically, the recurrent module only runs 2 iterations during gradient computation in attack and defense, allowing us to obtain useful gradients that approximate the true ones while keeping the risk of exploding and vanishing gradients at minimum. At actual inference time, the number of updates is switched back to the default 12 iterations.  We include a  more detailed study in \cref{supp:raft}.

\noindent We found that using simple $L_1$ distance as the similarity metric in \cref{eq4} was sufficient to yield effective results. Details on the selection of similarity metrics can be found in \cref{supp:sim}.

\subsection{Reversing Adversarial Attacks}

\noindent {\bf Dense Attacks.} Projected Gradient Descent (PGD) is a commonly applied strong dense attack. It finds worst-case perturbations by randomly initializing in the given bound and iteratively updating the attack vector, projecting it back into the bound whenever it is exceeded. We view video clips as 4-D tensors in our experiments and deploy PGD with various $L_{\infty}$ bounds, corresponding to different attack strengths. We applied 20 steps for all PGD attacks.
We also evaluate against AutoAttack \cite{croce2020autoattack}, a state-of-the-art parameter-free dense attack combining different objectives. In our experiments, we applied AutoPGD with a combination of cross-entropy and Difference of Logits Ratio (DLR) loss. We found 10 iterations were sufficiently strong.
For our defense, we apply $K=20$ steps of iterations with step size $\eta=2$ and $L_\infty$ bound $\epsilon_r=12$. 

Quantitative results on defending dense attacks are reported in \cref{pgd}. For comparison, "random" is applying uniform noise with the same bound as the attack/defense. Our defense is effective under various types and strengths of attacks. Clean accuracy is also preserved to a considerable degree. We show visualizations in \cref{fig:dense}. Dense attacks in the RGB space result in corrupted optical flow estimations and misclassified labels. After our defense, accurate optical flow is restored and classification results are corrected.

\begin{figure}[t]
 \centering
  \includegraphics[width=0.47 \textwidth]{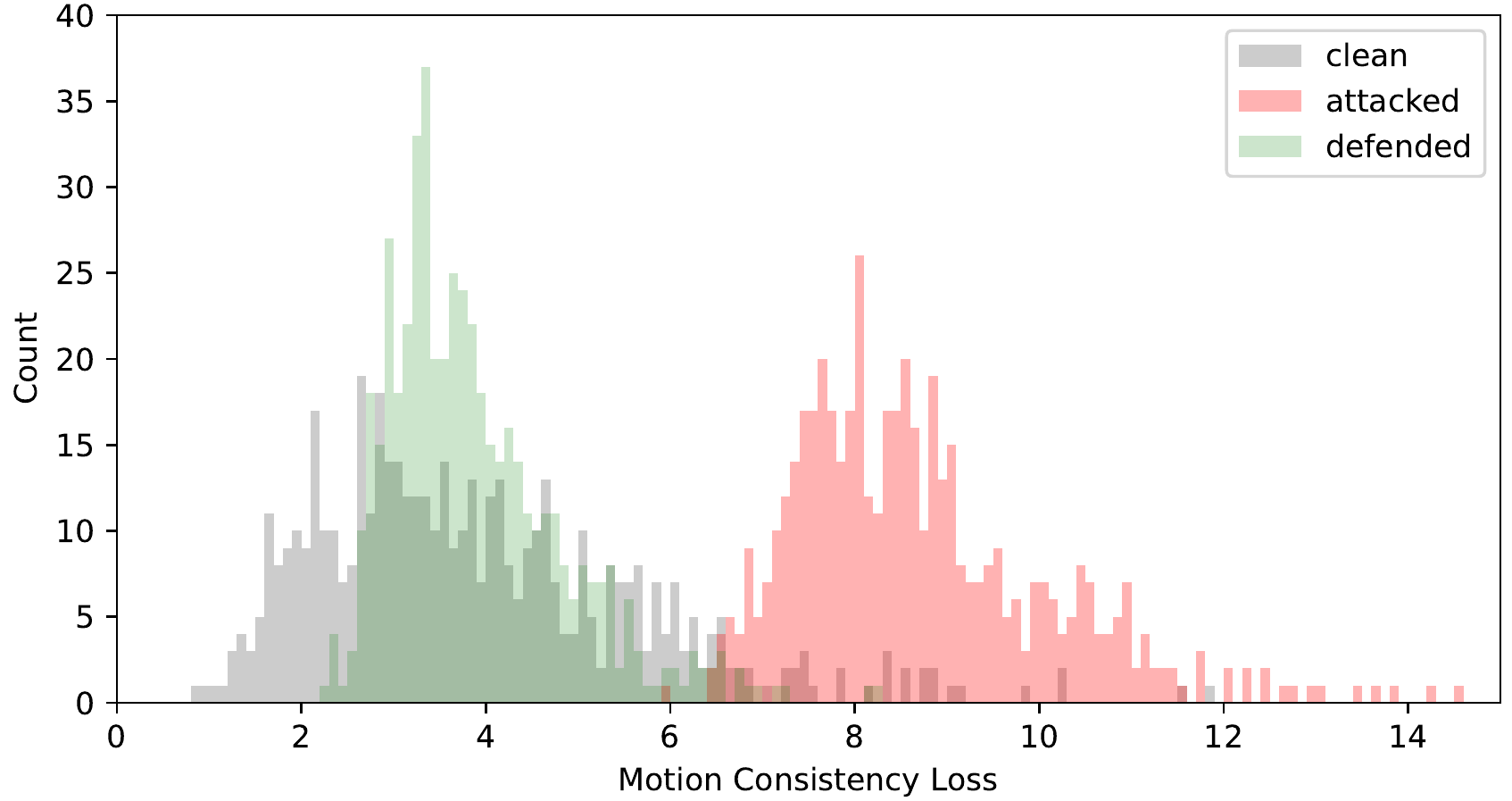}
  \caption{{\bf Motion Consistency Loss distribution.} The x axis is the Motion Consistency loss value, defined in Equation~\ref{eq4}, the y axis is frequency. Attacked videos (red) have significantly higher losses than clean ones (gray), whereas the defended videos (green) are pushed back to a lower loss distribution.}
  \label{loss}
\end{figure}

We can further plot the distribution of motion consistency losses over the test set videos for different examples. As shown in \cref{loss}, clean videos have relatively low losses, indicating that optical flow estimated from natural videos are consistent. Attacked videos have significantly higher losses, suggesting the motion consistency has been broken. Defended videos have lower losses, showing motion consistency has been restored.

\noindent {\bf One-Frame Attacks.} The one-frame attack introduced in \cite{Hwang2021JustOM} suggests that an attacker can identify the 'weak' frame in a video clip and add perturbations only to that frame to achieve success in fooling the model. While PGD attacks can be seen as the worst-case perturbation for an $L_p$ constraint, one-frame attacks are weaker since additional constraints are added to the attacker, trading off attack strength for imperceptibility.

 We follow the original paper's evaluation protocol of using 32-frame clips, as we found that one-frame attacks were much less effective on 64-frame clips. Results in \cref{tab: oneframe} show the effectiveness of our defense. We also find that one-frame attacks are fragile as the fooling rate can be greatly reduced by simply adding uniform noise on attacked videos. Moreover, we observe a natural resistance toward one-frame attacks in our model even without defense. 
 

\begin{table}
\centering
\small
    \begin{tabular}{c|c|c|c}
         \toprule
         & \multicolumn{3}{c}{UCF-101} \\
          Model   & Standard & Random & Defended    \\
         \midrule 
         Clean & 79.7 & 71.4 & 77.5 \\
         One-frame & 34.3 & 59.1 & {\bf 76.6} \\
         \bottomrule
    \end{tabular}
    \caption{\small{Adversarial robust accuracies against one-frame attacks on UCF-101. Our defense improves robustness by up to 42 points. 
}}

\label{tab: oneframe}
\end{table}

\begin{table}
\centering
\small
    \begin{tabular}{c|c|c}
         \toprule
         & \multicolumn{2}{c}{UCF-101} \\
          Model   & Standard  & Defended    \\
         \midrule 
         Clean & 84.0 & 84.0 \\
         Flickering & 34.0 & {\bf 79.5} \\ 
         \bottomrule
    \end{tabular}
    \caption{\small{Adversarial robust accuracies against flickering attacks on UCF-101. Our defense improves robustness by up to 45 points.
}}
 \vspace{-4mm}
\label{tab: flickering}
\end{table}

\noindent {\bf Over-the-air Flickering Attacks.} The authors of \cite{Pony2021OvertheAirAF} described an attack that only applies a constant color shift to each individual frame of a video. It is not constrained to an $L_p$ bound and aims at being imperceptible to human eyes by regularizing the magnitude and frequency of change (thickness and roughness) of the adversarial perturbations.
We follow the original paper and sample 90-frame clips. Flickering attacks are computationally expensive as an attack on each clip takes 1000 iterations. To reduce computational cost, we only perform 300 steps. Note that this makes the attack harder to defend as the perturbation has larger magnitudes, often exceeding our defense bound. Nevertheless, our defense remains effective. For flickering attacks, we evaluated on a 200-clip subset.

\begin{figure}[t]
 \centering
  \includegraphics[width=0.47\textwidth]{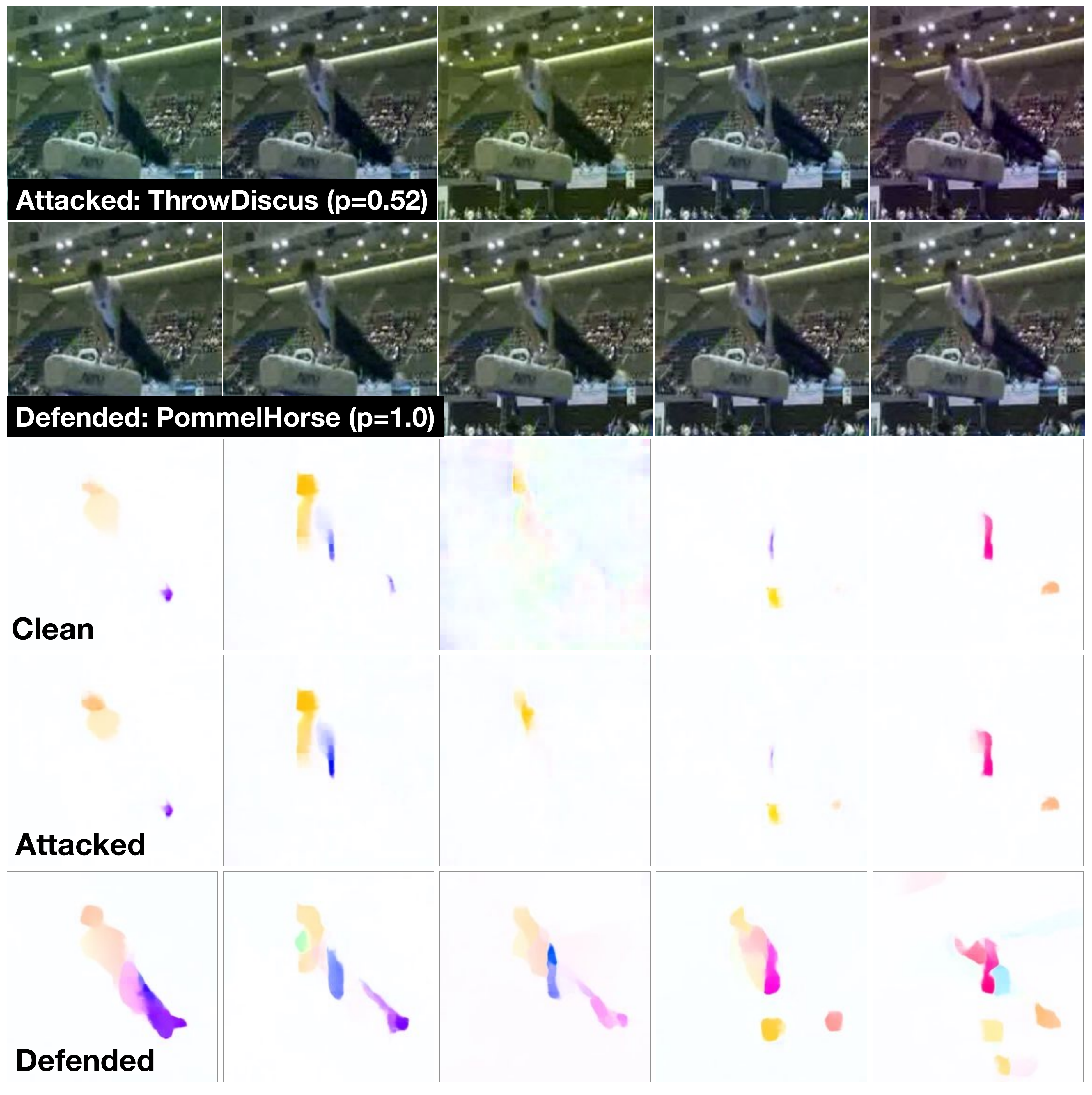}
  \caption{{\bf Resisting Flickering Attacks.} Flickering attacks fool the video action classifier by shifting the color, which is beyond the $L_{\infty}$ bounded attack. Row 1 is the RGB frames of a "PommelHorse" video under flickering attack. Row 2 is the defended frames. Row 3, 4, 5 are the flow fields estimated on clean, attacked and defended inputs, respectively. While flickering attacks do not significantly distort estimated flows, our defense is still effective because it suppresses the color shift in the inputs to lower consistency losses, which counters the adversarial attack.}
  \vspace{-5mm}
  \label{flicker}
\end{figure}

Interestingly, as shown in \cref{flicker}, we find that the flow fields were not significantly distorted. This is because the relative pixel values are not changed and pattern correspondence is still easy to find. However, the motion consistency loss is still much larger than in clean videos since there are shifts in pixel values between frames. As illustrated in \cref{fig:twofold}, even if an accurate flow learns to warp perfectly, there will still be a constant color difference between the true first frame and the reconstructed one. To minimize this difference, our defense learns to suppress the color shift, resulting in defended videos. In this case, the defended flows even exhibit better qualities than clean ones. 




 \vspace{-2mm}
\subsection{Defending Against Adaptive Attacks}

Evaluation of our defense against three adaptive attacks is summarized in \cref{tab: adaptive}. All attacks are $L_\infty=8$ bounded and defenses are $L_\infty=12$ bounded. All attacks and defenses are obtained through 20 iterations. Adaptive attacks are generally computationally intensive, experiments in this section are evaluated on a 200-clip subset of the test set. We can see that compared to static attacks, adaptive attacks are stronger as the defended accuracies are lowered to various extents. Notably, BPDA is the strongest attack. 

We also observe that our multiple constraint defense consistently improved robustness. This is because when multiple constraints are introduced, the attacker has to solve for a multi-task optimization, which is inherently hard due to more constraints. The attack budget has to be spent not only on fooling the classifier but also on satisfying the various constraints. Another consequence of this is weaker effectiveness of the attacks on standard models. The same lose-lose situation for the attacker described in \cite{Mao_2021_ICCV} is observed here, namely if the attacker chooses to take our defense into consideration, the effectiveness on undefended models is weakened; if not, our defense continues to be strongly effective.

\begin{table}[t]
\centering
\footnotesize
    \begin{tabular}{c|c|c|c}
         \toprule
         & \multicolumn{3}{c}{UCF-101} \\
          Model   & Standard & Defended & Defended (Multi)   \\
         \midrule 
         Clean & 85.5 & 83.5 & 81.0\\
         PGD$^*$ & 10.5 & 73.9 & - \\
         Adaptive I & 28.0 & 65.0 & {\bf 68.5} \\
         Adaptive II (FlowReg) & 64.5 & 68.5 & {\bf 72.5}\\
         Adaptive III (BPDA) & 18.5 & 20.5 & {\bf25.5}\\
         \bottomrule
    \end{tabular}
\caption{\small{Adversarial robust accuracies under adaptive attacks on UCF-101 dataset. We show robust accuracy under three types of adaptive attacks. For reference, we also show the standard PGD attack, where the * indicates that the attack is evaluated on 1000 examples. By applying our defense with multiple motion constraints, we can obtain a robust accuracy of 25.5, which is over 15 points higher than the standard defense attacked by PGD. Our defense also improves BPDA attacked videos by 7 points. Additionally, using multiple constraints consistently improves robustness.
}}
\vspace{-3mm}
\label{tab: adaptive}
\end{table}

         

\section{Conclusions}
In this work we show that adversarial attacks for video classification not only fool the classifier but also disrupt the motion consistency of the videos. We proposed a novel inference-time defense, improving robustness by restoring motion via self-supervised flow consistency. Empirical results show robustness under a variety of attacks. Our work is the first inference-time defense for videos that uses motion consistency for countering adversarial attacks. Our defense is robust even under strong adaptive attacks. Our work suggests that video data is hard to defend for existing methods because they ignore the rich motion structure in the natural data.

\section{Acknowledgements}

This research is based on work partially supported by the DARPA SAIL-ON program and the NSF NRI award \#1925157. This work was supported in part by a GE/DARPA grant, a CAIT grant, and gifts from JP Morgan, DiDi, and Accenture. We thank Mandi Zhao for discussion.

{\small
\bibliographystyle{ieee_fullname}
\bibliography{egbib}

\begin{thebibliography}{10}\itemsep=-1pt

\bibitem{tsipras2018robustness}


\bibitem{athalye2018obfuscated}
Anish Athalye, Nicholas Carlini, and David Wagner.
\newblock Obfuscated gradients give a false sense of security: Circumventing
  defenses to adversarial examples.
\newblock In {\em International conference on machine learning}, pages
  274--283. PMLR, 2018.

\bibitem{bengio1994vanishing}
Yoshua Bengio, Patrice Simard, and Paolo Frasconi.
\newblock Learning long-term dependencies with gradient descent is difficult.
\newblock {\em IEEE transactions on neural networks}, 5(2):157--166, 1994.

\bibitem{brox2004highacc}
Thomas Brox, Andr{\'e}s Bruhn, Nils Papenberg, and Joachim Weickert.
\newblock High accuracy optical flow estimation based on a theory for warping.
\newblock In {\em European conference on computer vision}, pages 25--36.
  Springer, 2004.

\bibitem{butler2012sintel}
Daniel~J Butler, Jonas Wulff, Garrett~B Stanley, and Michael~J Black.
\newblock A naturalistic open source movie for optical flow evaluation.
\newblock In {\em European conference on computer vision}, pages 611--625.
  Springer, 2012.

\bibitem{Carlini2017TowardsET}
Nicholas Carlini and David~A. Wagner.
\newblock Towards evaluating the robustness of neural networks.
\newblock {\em 2017 IEEE Symposium on Security and Privacy (SP)}, pages 39--57,
  2017.

\bibitem{carreira2017i3d}
Joao Carreira and Andrew Zisserman.
\newblock Quo vadis, action recognition? a new model and the kinetics dataset.
\newblock In {\em proceedings of the IEEE Conference on Computer Vision and
  Pattern Recognition}, pages 6299--6308, 2017.

\bibitem{attackcars}
Alesia Chernikova, Alina Oprea, Cristina Nita-Rotaru, and BaekGyu Kim.
\newblock Are self-driving cars secure? evasion attacks against deep neural
  networks for steering angle prediction.
\newblock In {\em 2019 IEEE Security and Privacy Workshops (SPW)}, pages
  132--137. IEEE, 2019.

\bibitem{cho2014gru}
Kyunghyun Cho, Bart Van~Merri{\"e}nboer, Dzmitry Bahdanau, and Yoshua Bengio.
\newblock On the properties of neural machine translation: Encoder-decoder
  approaches.
\newblock {\em arXiv preprint arXiv:1409.1259}, 2014.

\bibitem{cohen2019certified}
Jeremy Cohen, Elan Rosenfeld, and Zico Kolter.
\newblock Certified adversarial robustness via randomized smoothing.
\newblock In {\em International Conference on Machine Learning}, pages
  1310--1320. PMLR, 2019.

\bibitem{croce2022evaluating}
Francesco Croce, Sven Gowal, Thomas Brunner, Evan Shelhamer, Matthias Hein, and
  Taylan Cemgil.
\newblock Evaluating the adversarial robustness of adaptive test-time defenses.
\newblock {\em arXiv preprint arXiv:2202.13711}, 2022.

\bibitem{croce2020autoattack}
Francesco Croce and Matthias Hein.
\newblock Reliable evaluation of adversarial robustness with an ensemble of
  diverse parameter-free attacks.
\newblock In {\em International conference on machine learning}, pages
  2206--2216. PMLR, 2020.

\bibitem{attackface}
Yinpeng Dong, Hang Su, Baoyuan Wu, Zhifeng Li, Wei Liu, Tong Zhang, and Jun
  Zhu.
\newblock Efficient decision-based black-box adversarial attacks on face
  recognition.
\newblock In {\em Proceedings of the IEEE/CVF Conference on Computer Vision and
  Pattern Recognition}, pages 7714--7722, 2019.

\bibitem{dosovitskiy2015flownet}
Alexey Dosovitskiy, Philipp Fischer, Eddy Ilg, Philip Hausser, Caner Hazirbas,
  Vladimir Golkov, Patrick Van Der~Smagt, Daniel Cremers, and Thomas Brox.
\newblock Flownet: Learning optical flow with convolutional networks.
\newblock In {\em Proceedings of the IEEE international conference on computer
  vision}, pages 2758--2766, 2015.

\bibitem{feichtenhofer2016convolutional}
Christoph Feichtenhofer, Axel Pinz, and Andrew Zisserman.
\newblock Convolutional two-stream network fusion for video action recognition.
\newblock In {\em Proceedings of the IEEE conference on computer vision and
  pattern recognition}, pages 1933--1941, 2016.

\bibitem{Goodfellow2015ExplainingAH}
Ian~J. Goodfellow, Jonathon Shlens, and Christian Szegedy.
\newblock Explaining and harnessing adversarial examples.
\newblock {\em CoRR}, abs/1412.6572, 2015.

\bibitem{maskrcnn}
Kaiming He, Georgia Gkioxari, Piotr Doll{\'a}r, and Ross Girshick.
\newblock Mask r-cnn.
\newblock In {\em Proceedings of the IEEE international conference on computer
  vision}, pages 2961--2969, 2017.

\bibitem{horn1981determining}
Berthold~KP Horn and Brian~G Schunck.
\newblock Determining optical flow.
\newblock {\em Artificial intelligence}, 17(1-3):185--203, 1981.

\bibitem{Hwang2021JustOM}
Jaehui Hwang, Jun-Hyuk Kim, Jun-Ho Choi, and Jong-Seok Lee.
\newblock Just one moment: Structural vulnerability of deep action recognition
  against one frame attack.
\newblock {\em 2021 IEEE/CVF International Conference on Computer Vision
  (ICCV)}, pages 7648--7656, 2021.

\bibitem{flownet2}
Eddy Ilg, Nikolaus Mayer, Tonmoy Saikia, Margret Keuper, Alexey Dosovitskiy,
  and Thomas Brox.
\newblock Flownet 2.0: Evolution of optical flow estimation with deep networks.
\newblock In {\em Proceedings of the IEEE conference on computer vision and
  pattern recognition}, pages 2462--2470, 2017.

\bibitem{Inkawhich2018AdversarialForOptical}
Nathan Inkawhich, Matthew~J. Inkawhich, Yiran Chen, and Hai~Helen Li.
\newblock Adversarial attacks for optical flow-based action recognition
  classifiers.
\newblock {\em ArXiv}, abs/1811.11875, 2018.

\bibitem{3dcnn}
Shuiwang Ji, Wei Xu, Ming Yang, and Kai Yu.
\newblock 3d convolutional neural networks for human action recognition.
\newblock {\em IEEE transactions on pattern analysis and machine intelligence},
  35(1):221--231, 2012.

\bibitem{Jia2019IdentifyingAR}
Xiaojun Jia, Xingxing Wei, and Xiaochun Cao.
\newblock Identifying and resisting adversarial videos using temporal
  consistency.
\newblock {\em ArXiv}, abs/1909.04837, 2019.

\bibitem{jonschkowski2020matters}
Rico Jonschkowski, Austin Stone, Jonathan~T Barron, Ariel Gordon, Kurt
  Konolige, and Anelia Angelova.
\newblock What matters in unsupervised optical flow.
\newblock In {\em European Conference on Computer Vision}, pages 557--572.
  Springer, 2020.

\bibitem{kang2021stable}
Qiyu Kang, Yang Song, Qinxu Ding, and Wee~Peng Tay.
\newblock Stable neural ode with lyapunov-stable equilibrium points for
  defending against adversarial attacks.
\newblock {\em Advances in Neural Information Processing Systems},
  34:14925--14937, 2021.

\bibitem{kinfu2022advtrainvid}
Kaleab~A Kinfu and Ren{\'e} Vidal.
\newblock Analysis and extensions of adversarial training for video
  classification.
\newblock In {\em Proceedings of the IEEE/CVF Conference on Computer Vision and
  Pattern Recognition}, pages 3416--3425, 2022.

\bibitem{koffka1922perception}
Kurt Koffka.
\newblock Perception: an introduction to the gestalt-theorie.
\newblock {\em Psychological bulletin}, 19(10):531, 1922.

\bibitem{alexnet}
Alex Krizhevsky, Ilya Sutskever, and Geoffrey~E Hinton.
\newblock Imagenet classification with deep convolutional neural networks.
\newblock In F. Pereira, C.J. Burges, L. Bottou, and K.Q. Weinberger, editors,
  {\em Advances in Neural Information Processing Systems}, volume~25. Curran
  Associates, Inc., 2012.

\bibitem{kuehne2011hmdb51}
Hildegard Kuehne, Hueihan Jhuang, Est{\'\i}baliz Garrote, Tomaso Poggio, and
  Thomas Serre.
\newblock Hmdb: a large video database for human motion recognition.
\newblock In {\em 2011 International conference on computer vision}, pages
  2556--2563. IEEE, 2011.

\bibitem{kurakin2017atscale}
Alexey Kurakin, Ian~J. Goodfellow, and Samy Bengio.
\newblock Adversarial machine learning at scale.
\newblock In {\em International Conference on Learning Representations}, 2017.

\bibitem{advphysical}
Alexey Kurakin, Ian~J Goodfellow, and Samy Bengio.
\newblock Adversarial examples in the physical world.
\newblock In {\em Artificial intelligence safety and security}, pages 99--112.
  Chapman and Hall/CRC, 2018.

\bibitem{lawhon2022using}
Matthew Lawhon, Chengzhi Mao, and Junfeng Yang.
\newblock Using multiple self-supervised tasks improves model robustness.
\newblock {\em arXiv preprint arXiv:2204.03714}, 2022.

\bibitem{liu2022landscape}
Ruoshi Liu, Chengzhi Mao, Purva Tendulkar, Hao Wang, and Carl Vondrick.
\newblock Landscape learning for neural network inversion.
\newblock {\em arXiv e-prints}, pages arXiv--2206, 2022.

\bibitem{lo2021unforeseenvideos}
Shao-Yuan Lo and Vishal~M Patel.
\newblock Defending against multiple and unforeseen adversarial videos.
\newblock {\em IEEE Transactions on Image Processing}, 31:962--973, 2021.

\bibitem{lucas1981iterative}
Bruce~D Lucas, Takeo Kanade, et~al.
\newblock {\em An iterative image registration technique with an application to
  stereo vision}, volume~81.
\newblock Vancouver, 1981.

\bibitem{madry2018towards}
Aleksander Madry, Aleksandar Makelov, Ludwig Schmidt, Dimitris Tsipras, and
  Adrian Vladu.
\newblock Towards deep learning models resistant to adversarial attacks.
\newblock In {\em International Conference on Learning Representations}, 2018.

\bibitem{Mao_2021_ICCV}
Chengzhi Mao, Mia Chiquier, Hao Wang, Junfeng Yang, and Carl Vondrick.
\newblock Adversarial attacks are reversible with natural supervision.
\newblock In {\em Proceedings of the IEEE/CVF International Conference on
  Computer Vision (ICCV)}, pages 661--671, October 2021.

\bibitem{mao2020multitask}
Chengzhi Mao, Amogh Gupta, Vikram Nitin, Baishakhi Ray, Shuran Song, Junfeng
  Yang, and Carl Vondrick.
\newblock Multitask learning strengthens adversarial robustness.
\newblock In {\em European Conference on Computer Vision}, pages 158--174.
  Springer, 2020.

\bibitem{mao2022robust}
Chengzhi Mao, Lingyu Zhang, Abhishek Joshi, Junfeng Yang, Hao Wang, and Carl
  Vondrick.
\newblock Robust perception through equivariance, 2022.

\bibitem{mao19metric}
Chengzhi Mao, Ziyuan Zhong, Junfeng Yang, Carl Vondrick, and Baishakhi Ray.
\newblock Metric learning for adversarial robustness.
\newblock In {\em Advances in Neural Information Processing Systems},
  volume~32. Curran Associates, Inc., 2019.

\bibitem{memin1998denseOF}
Etienne M{\'e}min and Patrick P{\'e}rez.
\newblock Dense estimation and object-based segmentation of the optical flow
  with robust techniques.
\newblock {\em IEEE Transactions on Image Processing}, 7(5):703--719, 1998.

\bibitem{ng2018actionflownet}
Joe Yue-Hei Ng, Jonghyun Choi, Jan Neumann, and Larry~S Davis.
\newblock Actionflownet: Learning motion representation for action recognition.
\newblock In {\em 2018 IEEE Winter Conference on Applications of Computer
  Vision (WACV)}, pages 1616--1624. IEEE, 2018.

\bibitem{nie2022diffusion}
Weili Nie, Brandon Guo, Yujia Huang, Chaowei Xiao, Arash Vahdat, and Anima
  Anandkumar.
\newblock Diffusion models for adversarial purification.
\newblock {\em arXiv preprint arXiv:2205.07460}, 2022.

\bibitem{plizzari2022egomotion}
Chiara Plizzari, Mirco Planamente, Gabriele Goletto, Marco Cannici, Emanuele
  Gusso, Matteo Matteucci, and Barbara Caputo.
\newblock E2 (go) motion: Motion augmented event stream for egocentric action
  recognition.
\newblock In {\em Proceedings of the IEEE/CVF Conference on Computer Vision and
  Pattern Recognition}, pages 19935--19947, 2022.

\bibitem{Pony2021OvertheAirAF}
Roi Pony, Itay Naeh, and Shie Mannor.
\newblock Over-the-air adversarial flickering attacks against video recognition
  networks.
\newblock {\em 2021 IEEE/CVF Conference on Computer Vision and Pattern
  Recognition (CVPR)}, pages 515--524, 2021.

\bibitem{sevilla2018integration}
Laura Sevilla-Lara, Yiyi Liao, Fatma G{\"u}ney, Varun Jampani, Andreas Geiger,
  and Michael~J Black.
\newblock On the integration of optical flow and action recognition.
\newblock In {\em German conference on pattern recognition}, pages 281--297.
  Springer, 2018.

\bibitem{SimonGabriel2019FirstOrderAV}
Carl-Johann Simon-Gabriel, Yann Ollivier, L{\'e}on Bottou, Bernhard
  Sch{\"o}lkopf, and David Lopez-Paz.
\newblock First-order adversarial vulnerability of neural networks and input
  dimension.
\newblock In {\em ICML}, 2019.

\bibitem{simonyan2014twostream}
Karen Simonyan and Andrew Zisserman.
\newblock Two-stream convolutional networks for action recognition in videos.
\newblock {\em Advances in neural information processing systems}, 27, 2014.

\bibitem{vgg}
Karen Simonyan and Andrew Zisserman.
\newblock Very deep convolutional networks for large-scale image recognition.
\newblock {\em arXiv preprint arXiv:1409.1556}, 2014.

\bibitem{soomro2012ucf101}
Khurram Soomro, Amir~Roshan Zamir, and Mubarak Shah.
\newblock Ucf101: A dataset of 101 human actions classes from videos in the
  wild.
\newblock {\em arXiv preprint arXiv:1212.0402}, 2012.

\bibitem{sun2018pwc}
Deqing Sun, Xiaodong Yang, Ming-Yu Liu, and Jan Kautz.
\newblock Pwc-net: Cnns for optical flow using pyramid, warping, and cost
  volume.
\newblock In {\em Proceedings of the IEEE conference on computer vision and
  pattern recognition}, pages 8934--8943, 2018.

\bibitem{Szegedy2014IntriguingPO}
Christian Szegedy, Wojciech Zaremba, Ilya Sutskever, Joan Bruna, D. Erhan,
  Ian~J. Goodfellow, and Rob Fergus.
\newblock Intriguing properties of neural networks.
\newblock {\em CoRR}, abs/1312.6199, 2014.

\bibitem{convspatio-temp}
Graham~W Taylor, Rob Fergus, Yann LeCun, and Christoph Bregler.
\newblock Convolutional learning of spatio-temporal features.
\newblock In {\em European conference on computer vision}, pages 140--153.
  Springer, 2010.

\bibitem{teed2020raft}
Zachary Teed and Jia Deng.
\newblock Raft: Recurrent all-pairs field transforms for optical flow.
\newblock In {\em European conference on computer vision}, pages 402--419.
  Springer, 2020.

\bibitem{wang2019bidirectional}
Hao Wang, Chengzhi Mao, Hao He, Mingmin Zhao, Tommi~S Jaakkola, and Dina
  Katabi.
\newblock Bidirectional inference networks: A class of deep bayesian networks
  for health profiling.
\newblock In {\em Proceedings of the AAAI Conference on Artificial
  Intelligence}, volume~33, pages 766--773, 2019.

\bibitem{Wei2019SparseAP}
Xingxing Wei, Jun Zhu, and Hang Su.
\newblock Sparse adversarial perturbations for videos.
\newblock {\em ArXiv}, abs/1803.02536, 2019.

\bibitem{wei2020blackbox}
Zhipeng Wei, Jingjing Chen, Xingxing Wei, Linxi Jiang, Tat-Seng Chua, Fengfeng
  Zhou, and Yu-Gang Jiang.
\newblock Heuristic black-box adversarial attacks on video recognition models.
\newblock In {\em Proceedings of the AAAI Conference on Artificial
  Intelligence}, volume~34, pages 12338--12345, 2020.

\bibitem{wu2021attacking}
Boxi Wu, Heng Pan, Li Shen, Jindong Gu, Shuai Zhao, Zhifeng Li, Deng Cai,
  Xiaofei He, and Wei Liu.
\newblock Attacking adversarial attacks as a defense.
\newblock {\em arXiv preprint arXiv:2106.04938}, 2021.

\bibitem{Xiao2019AdvITAF}
Chaowei Xiao, Ruizhi Deng, Bo Li, Taesung Lee, Ben Edwards, Jinfeng Yi,
  Dawn~Xiaodong Song, Mingyan~D. Liu, and Ian Molloy.
\newblock Advit: Adversarial frames identifier based on temporal consistency in
  videos.
\newblock {\em 2019 IEEE/CVF International Conference on Computer Vision
  (ICCV)}, pages 3967--3976, 2019.

\bibitem{TV-L1}
Christopher Zach, Thomas Pock, and Horst Bischof.
\newblock A duality based approach for realtime tv-l1 optical flow.
\newblock In {\em DAGM-Symposium}, 2007.

\bibitem{zhang2019limitations}
Huan Zhang, Hongge Chen, Zhao Song, Duane Boning, Inderjit~S Dhillon, and
  Cho-Jui Hsieh.
\newblock The limitations of adversarial training and the blind-spot attack.
\newblock {\em arXiv preprint arXiv:1901.04684}, 2019.

\bibitem{zhang2019theoretically}
Hongyang Zhang, Yaodong Yu, Jiantao Jiao, Eric~P. Xing, Laurent~El Ghaoui, and
  Michael~I. Jordan.
\newblock Theoretically principled trade-off between robustness and accuracy.
\newblock In {\em International Conference on Machine Learning}, 2019.

\end{thebibliography}
}


\twocolumn[
\null
{\vspace*{-.3in}}{\vskip .375in}
\begin{center}
{{\Large \bf {Appendix} \par}}
  {\vspace*{-22pt}}{\vspace*{24pt}}
  {
  \large
  \lineskip .5em
  \par
  }
  \vskip .5em
  \vspace*{12pt}
\end{center}
]

\thispagestyle{empty}

\appendix

\newpage

\section{Results}
\subsection{Raft Recurrent Steps}
\label{supp:raft}
\begin{figure}[h]
\centering
   \includegraphics[width=0.45\textwidth]{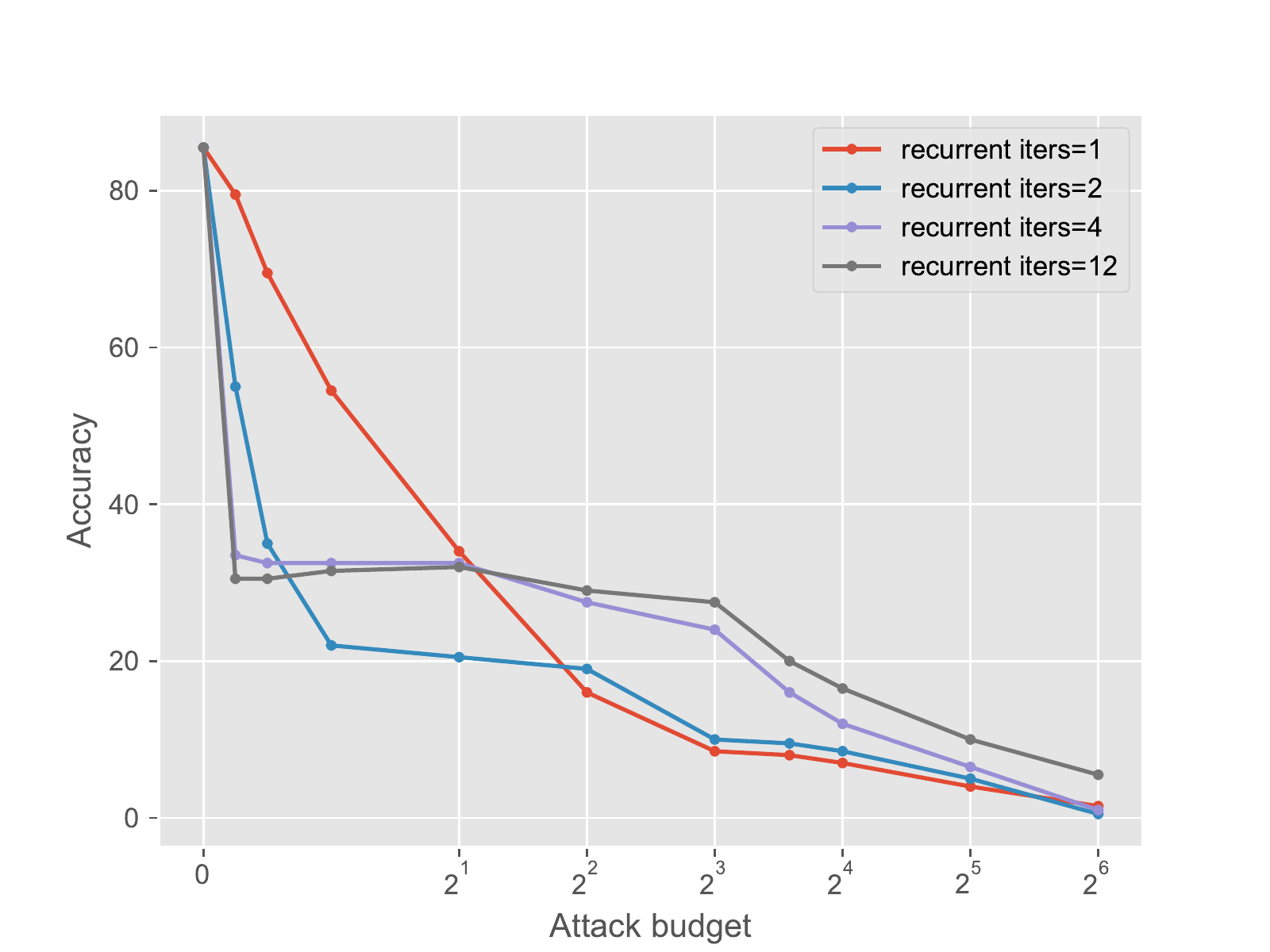}
  \caption{Accuracies under PGD attacks of RAFT using different iters in the recurrent module. The x-axis is the attack budget and the y-axis is the accuracy of models. Every curve corresponds to a model with a specific number of recurrent steps in the GRU module when computing gradients. Note that after the attack is generated, all models still perform full 12 recurrent steps at final inference. We find that for larger attack budgets, using less recurrent steps to approximate gradients results in reduced accuracy. If the number of recurrent steps is too small, the gradient approximation becomes inaccurate, also rendering attack ineffective. We balance this by choosing recurrent iters=2.} 
  \label{fig:raft}
\end{figure}

The GRU\cite{cho2014gru} module in RAFT \cite{teed2020raft} has a default setting of 12 recurrent steps. When directly applying this setting, we find that the undefended model has a natural robust accuracy of above 20\% under end-to-end $L_\infty=8$ PGD attacks. While this could be a benefit of optical flow's invariance to appearance, we also investigate the possibility of reliance on obfuscated gradients. It is well-known that recurrent neural networks are prone to vanishing and exploding gradients \cite{bengio1994vanishing}. We examine this possibility by approximating the gradients of the GRU during the attack. Specifically, we simply reduce  the number of recurrent steps when computing gradients, then perform full 12-step forward pass during final inference. Robust accuracies under various attack budgets using different recurrent iters are plotted in \cref{fig:raft}.

We find that for attacks with large budgets, the natural undefended accuracy can be reduced by approximating gradients, indicating a certain extent of reliance on obfuscated gradients when applying full 12 recurrent steps. However, when the number of steps is too small, the gradients are not an accurate approximation of the inference time 12-step gradients. We chose recurrent iters = 2 as it balances the two: remains a good approximation of the 12-step forward pass gradients while keeping the reliance on obfuscated gradients at minimum. This baseline has a reduced undefended accuracy of 10.5\%.

\subsection{Similarity Metrics}
\label{supp:sim}
In Section 3.2 of the paper, we described using a similarity metric between warped and ground truth frames as the photometric consistency loss. As the authors of \cite{jonschkowski2020matters} have discussed, there are many choices for this metric. We experimented on the $L_1$, $L_2$, and $SSIM$ distances. Robust accuracies using only photometric loss with different similarity metrics are reported in \cref{sim}. In terms of improving defense effectiveness, $L_1$ is 
better than or same as other metrics so for this work we used the simplest $L_1$ distance for photometric consistency.

\begin{table}[!h]
\centering
\small
    \begin{tabular}{c|c}
         \toprule
          metric   & Defended Acc  \\
         \midrule 
         $L_1$ & 66.5 \\
         $L_2$ & 61.5 \\
         SSIM & 66.5\\

         \bottomrule
    \end{tabular}
    
\caption{\small{Metrics
}}
\vspace{-3mm}
\label{sim}
\end{table}

\subsection{Hyperparameter of Adaptive Attacks}

In section 4.3 of our paper, we described our adaptive attacks. For Adaptive Attack I and Adaptive Attack II (FlowReg) we followed \cite{Mao_2021_ICCV}, applying the attack by solving a Lagrangian that incorporates a defend-aware term:
\begin{equation}
\begin{aligned}
    \delta_{{\bf AA1}} &= \argmax_{\delta} \mathcal{L_\text{CE}} - \lambda_1 \mathcal{L_{MC}}\\
    \delta_{{\bf AA2}} &= \argmax_{\delta} \mathcal{L_\text{CE}} - \lambda_2 |G(x+\delta) - G(x)|_p
\end{aligned}
\end{equation}

\begin{figure}[h]
\centering
   \includegraphics[width=0.4\textwidth]{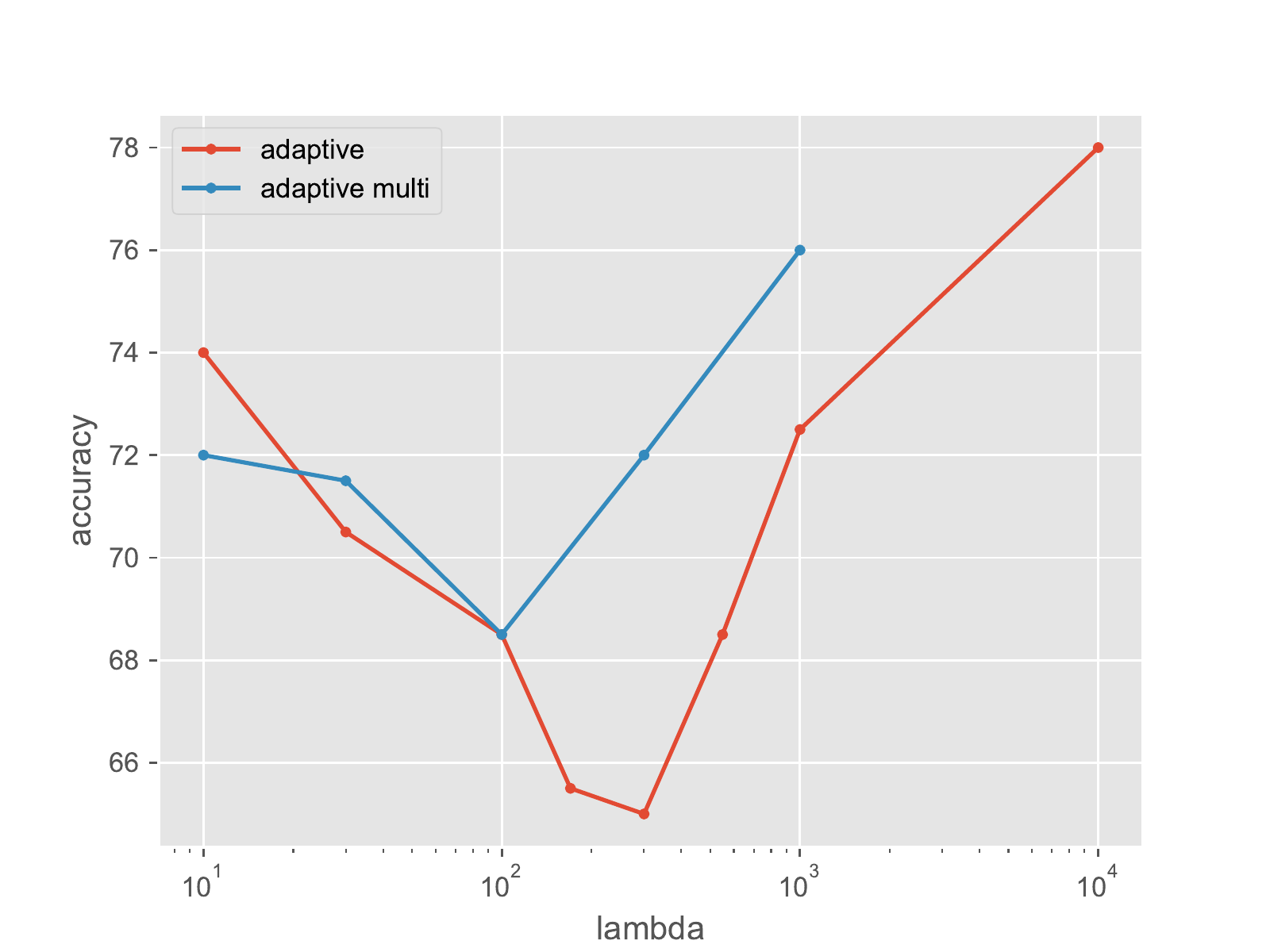}
   \includegraphics[width=0.4\textwidth]{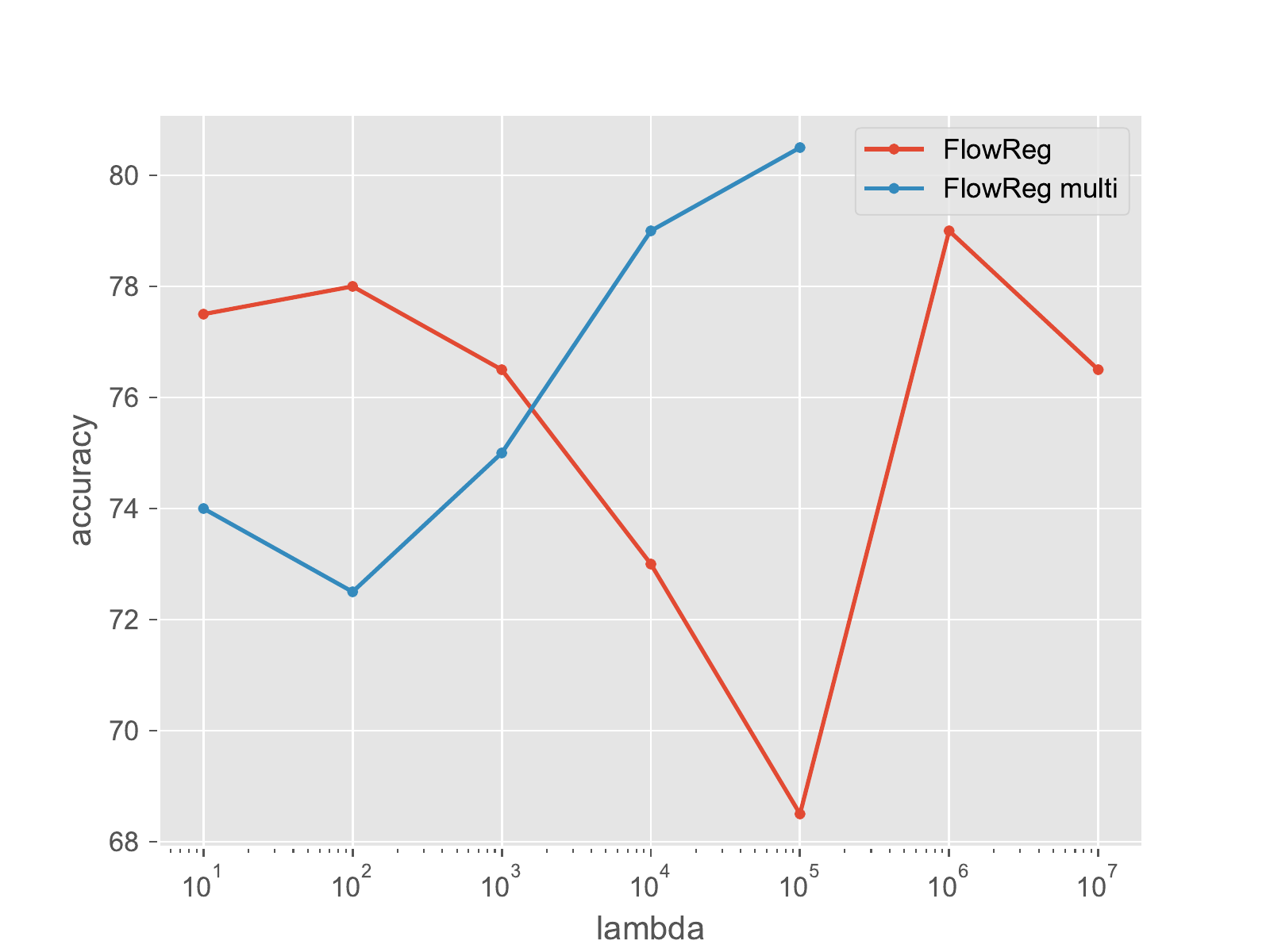}
  \caption{Parameter sweep for the strongest adaptive attacks. In both plots the x-axis is the value of $\lambda$, the y-axis is the defended accuracy of each adaptive attack. Lower is better, indicating that the attack is more capable of circumventing the defense. Red curves are using the baseline defense, blue curves are using multiple-constraints defense. The left figure is Adaptive Attack I, where the optimal $\lambda=300$. The right is Adaptive Attack II, where the optimal  $\lambda=10^5$}
  \label{adaptive}
\end{figure}

We perform a parameter sweep of $\lambda$ to find the strongest attacks. This is shown in \cref{adaptive}

\subsection{Defense Steps and Clip Length}

\begin{figure}[!h]
\centering
   \includegraphics[width=0.33\textwidth]{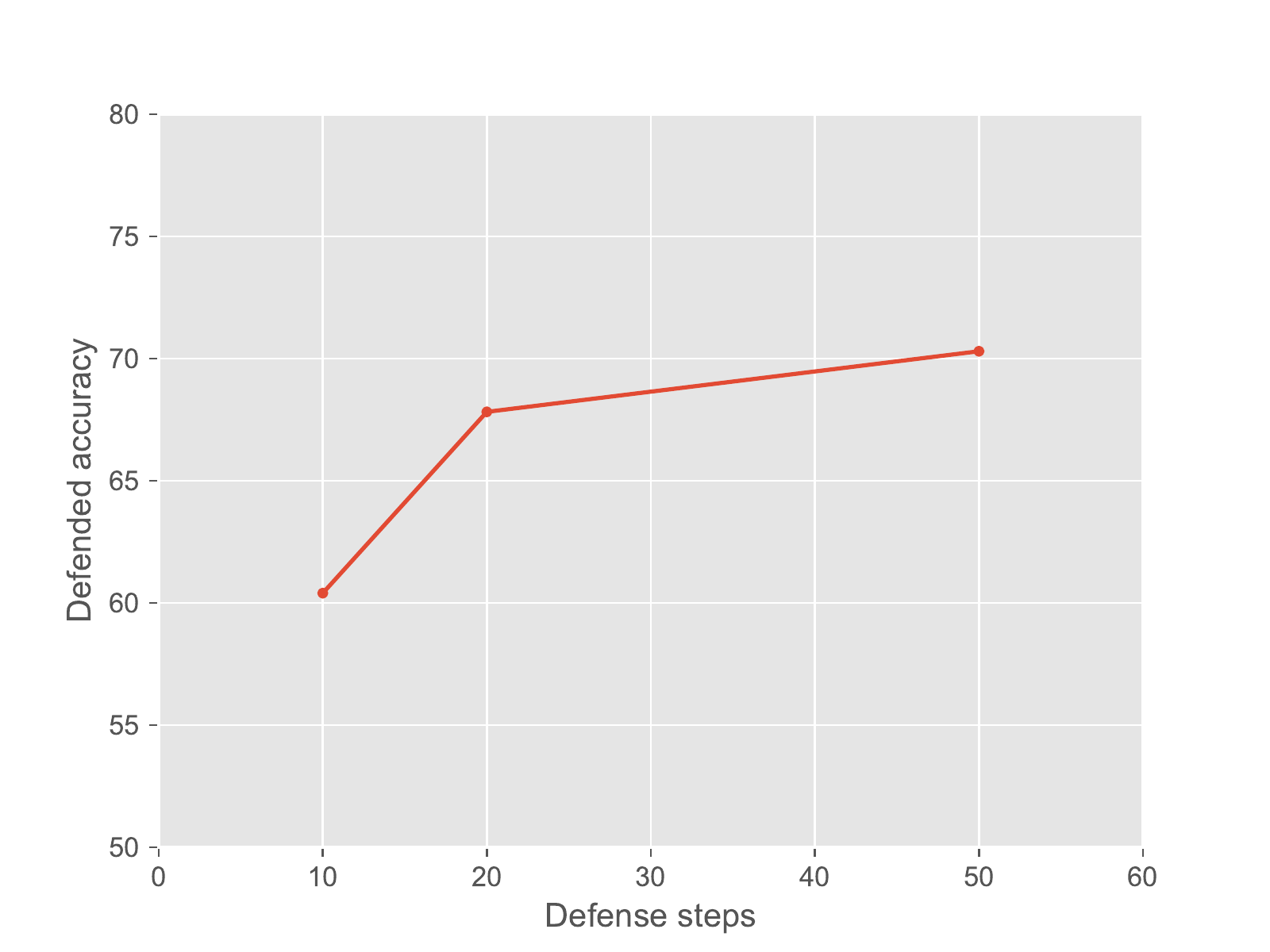}
   \includegraphics[width=0.33\textwidth]{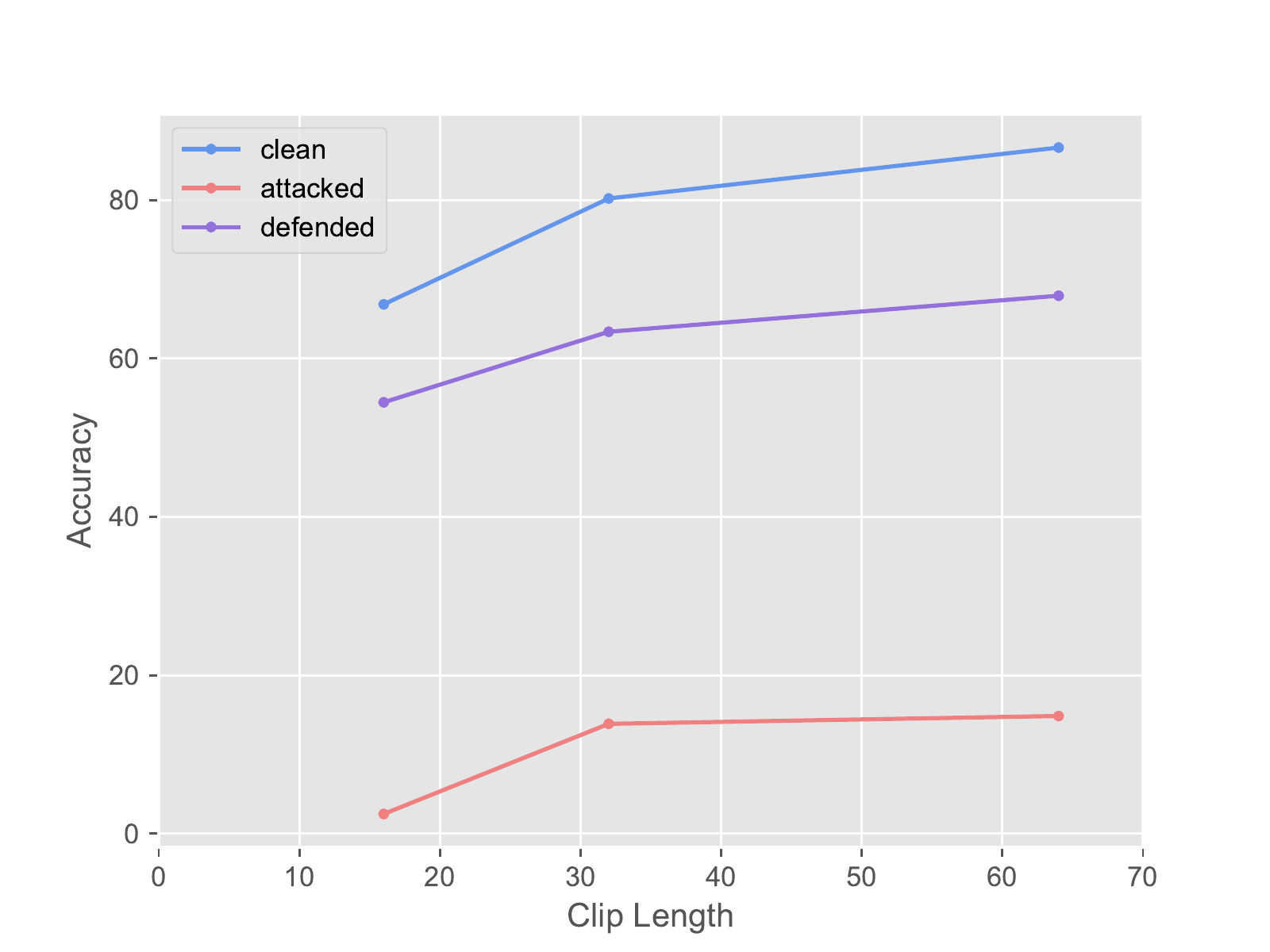}
   \includegraphics[width=0.33\textwidth]{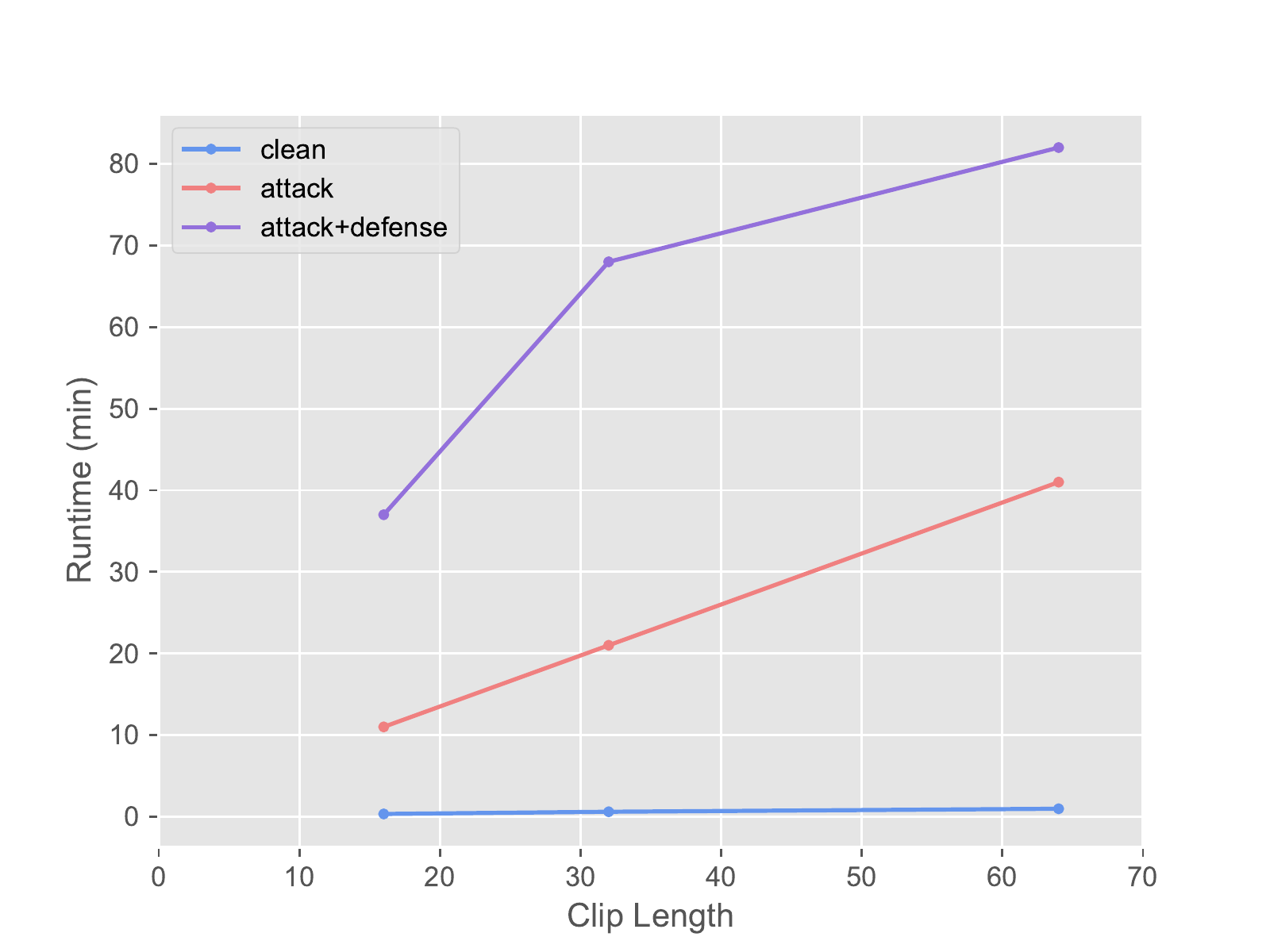}
  \caption{Effect of defense steps and clip Length. In the figure on the left, we show how defended accuracy is affected by the change in the number of defense steps. We find 20 steps to be sufficient for high effectiveness. In the figure in the middle, we show how the length of clips effect clean, attacked, and defended accuracies. In the figure on the right, we show how the length of clips affects the runtime of inference.}
  \label{def params}
\end{figure}

In \cref{def params}, we report the effect of defense steps and clip lengths. 

\section{Implementation Details}
\subsection{Efficient Implementation of RAFT for Videos}
The original RAFT \cite{teed2020raft} was implemented for computing optical flow between one pair of frames. They first use an encoder to extract features per-pixel. This is inefficient when computing stacks of flow fields for a video clip with multiple pairs of frames, as all frames except for the first and the last will be passed through the encoder twice, once as the source frame and once as the target frame. We modified the implementation by pre-computing  the visual features of every frame first, and using them when needed. This resulted in a faster and memory-efficient implementation for video clips.

\begin{table}[h]
\centering
\small
    \begin{tabular}{c|c|c}
         \toprule
             & Original & Efficient \\
         \midrule 
         Runtime (s) & 23.27 & {\bf 17.61}\\
         GPU Memory (MB)& 30,888 & {\bf 27,056}\\
         \bottomrule
    \end{tabular}
    
\caption{\small{Efficient implementation of RAFT for videos.
. Results are reported performing one forward pass of a 64-frame clip. Our efficient implementation reduces runtime and memory consumption.}}
\vspace{-3mm}
\label{sim}
\end{table}

\subsection{Other Motion Consistency Consctraints}
Unsupervised optical flow learning has a rich literature, providing us with many handy tricks for measuring optical flow quality. In our work, we only experimented with the basic photometric consistency and smoothness constraints. It is natural to extend our method to including more flow quality assessment techniques, for example, occlusion-aware consistency. We leave that to future work.

\section{Visualizations}

\begin{figure*}
\centering
  \vspace{-10mm}
   \includegraphics[width=0.9\textwidth]{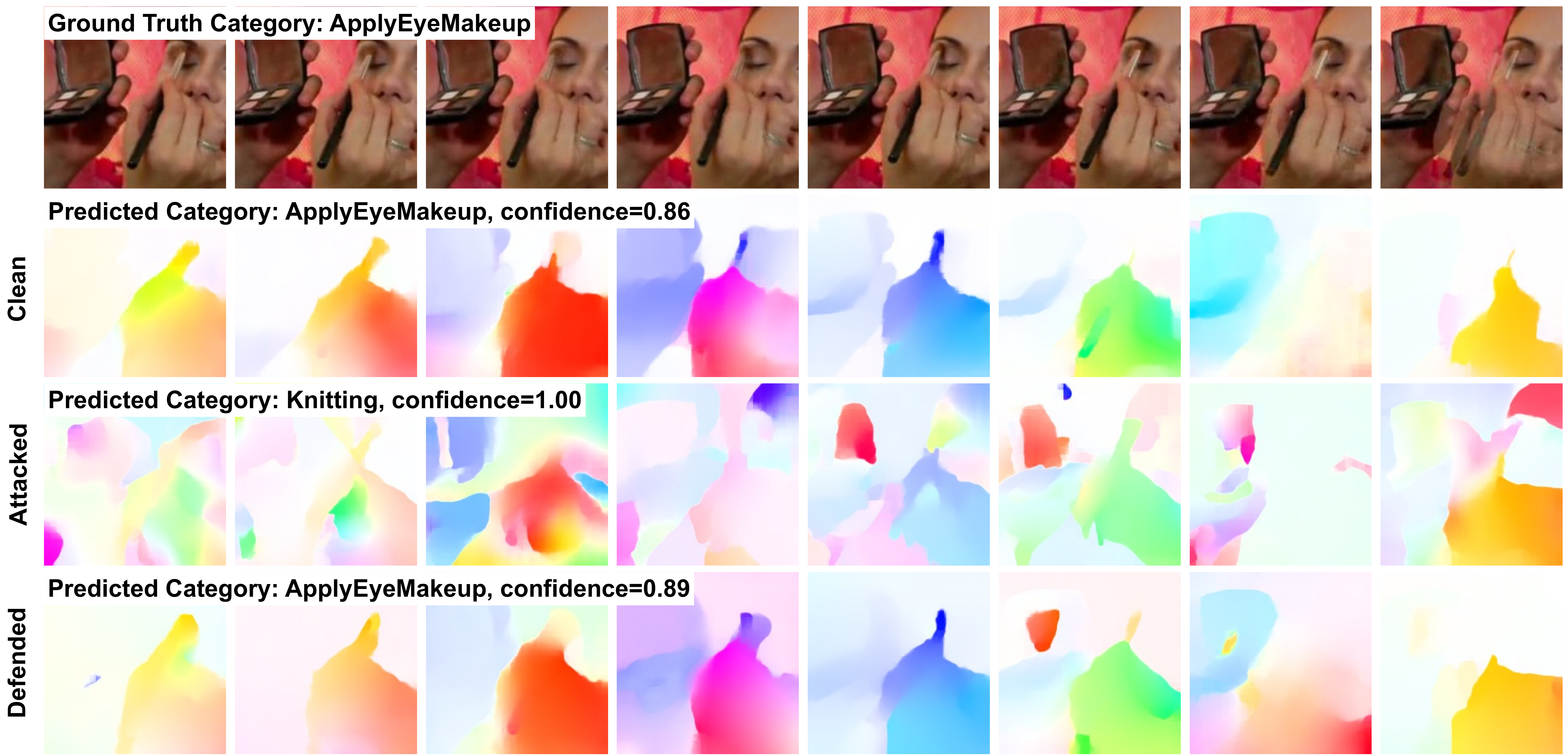}
  \vspace{-7mm}
 \label{fig:flpw}
\end{figure*}

\begin{figure*}[!h]
\centering
   \includegraphics[width=0.9\textwidth]{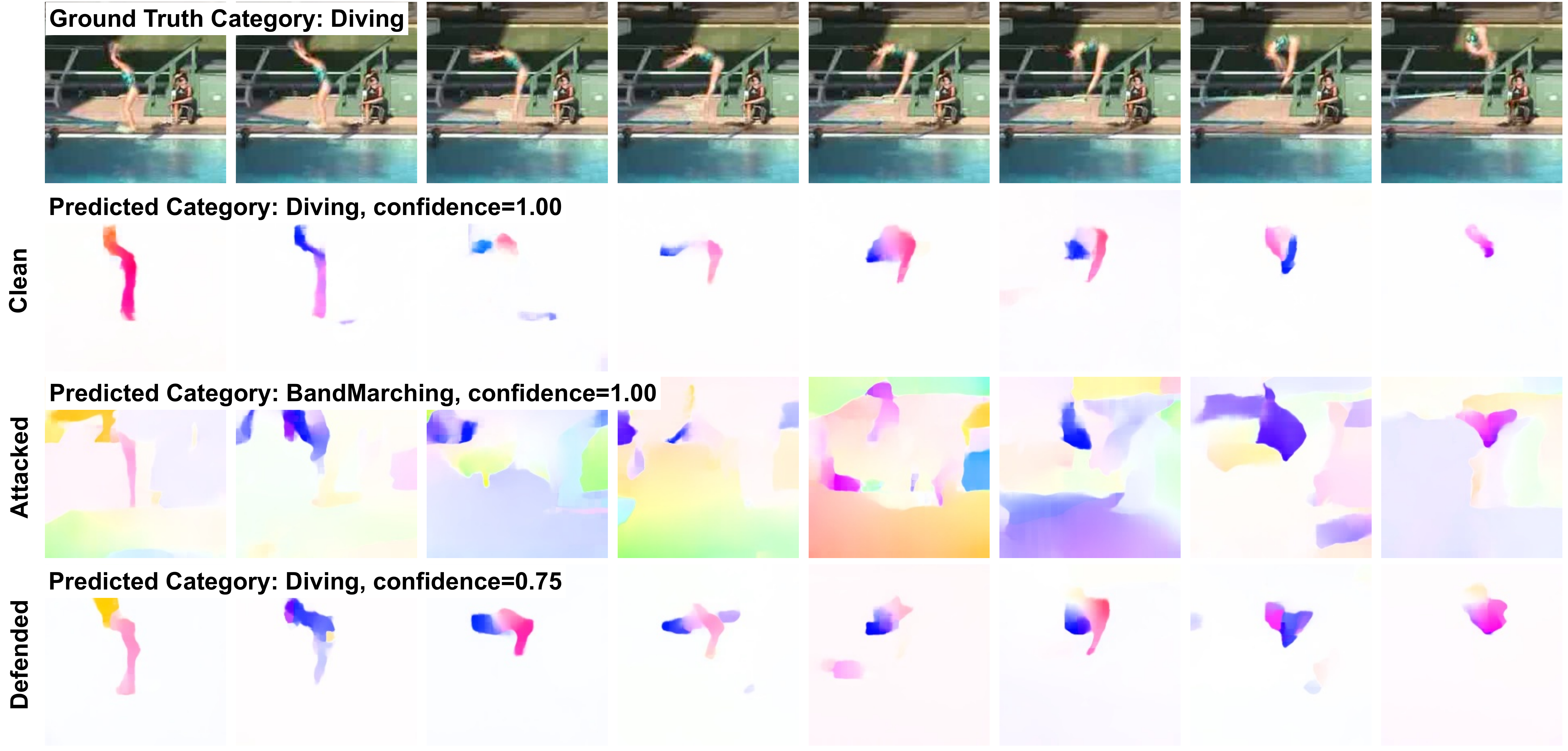}
  \label{fig:flpw}
   \vspace{-7mm}
\end{figure*}

\begin{figure*}[!h]
\centering
   \includegraphics[width=0.9\textwidth]{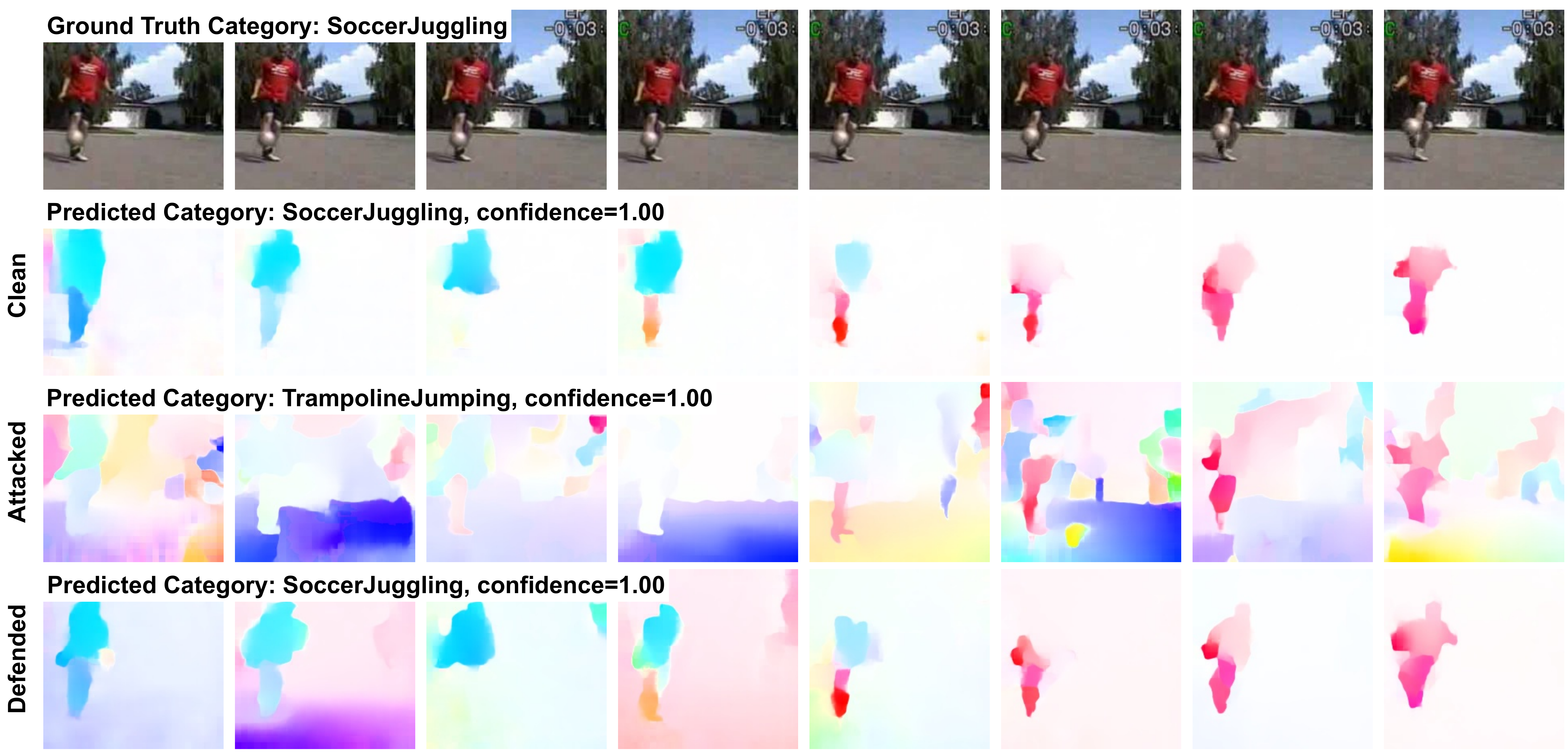}
  \label{fig:flpw}
  \vspace{-7mm}
\end{figure*}

\begin{figure*}[!h]
\vspace{-10mm}
\centering
 \includegraphics[width=0.9\textwidth]{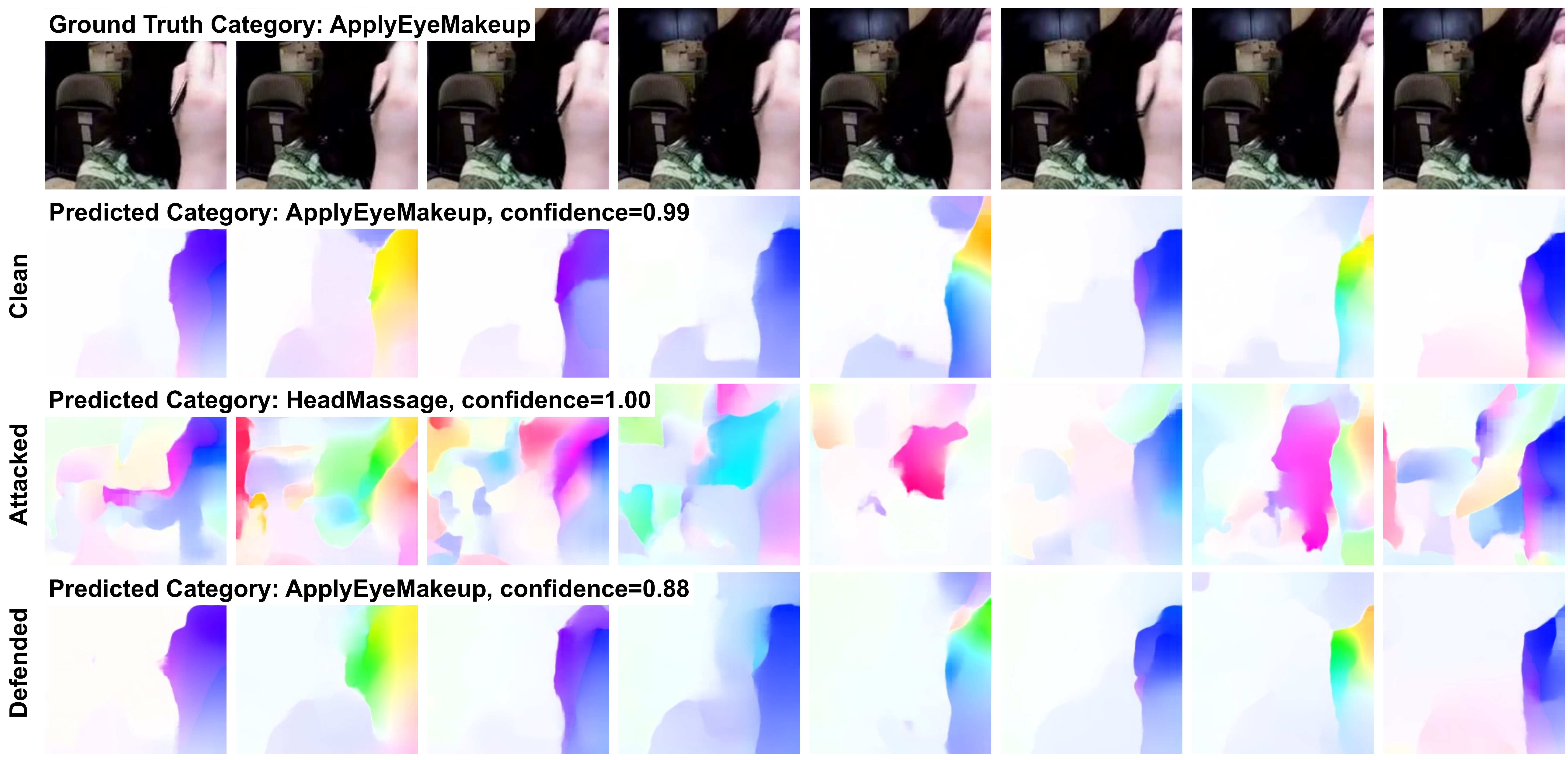}
  \label{fig:flpw}
  \vspace{-7mm}
\end{figure*}

\begin{figure*}[!h]
\centering
   \includegraphics[width=0.9\textwidth]{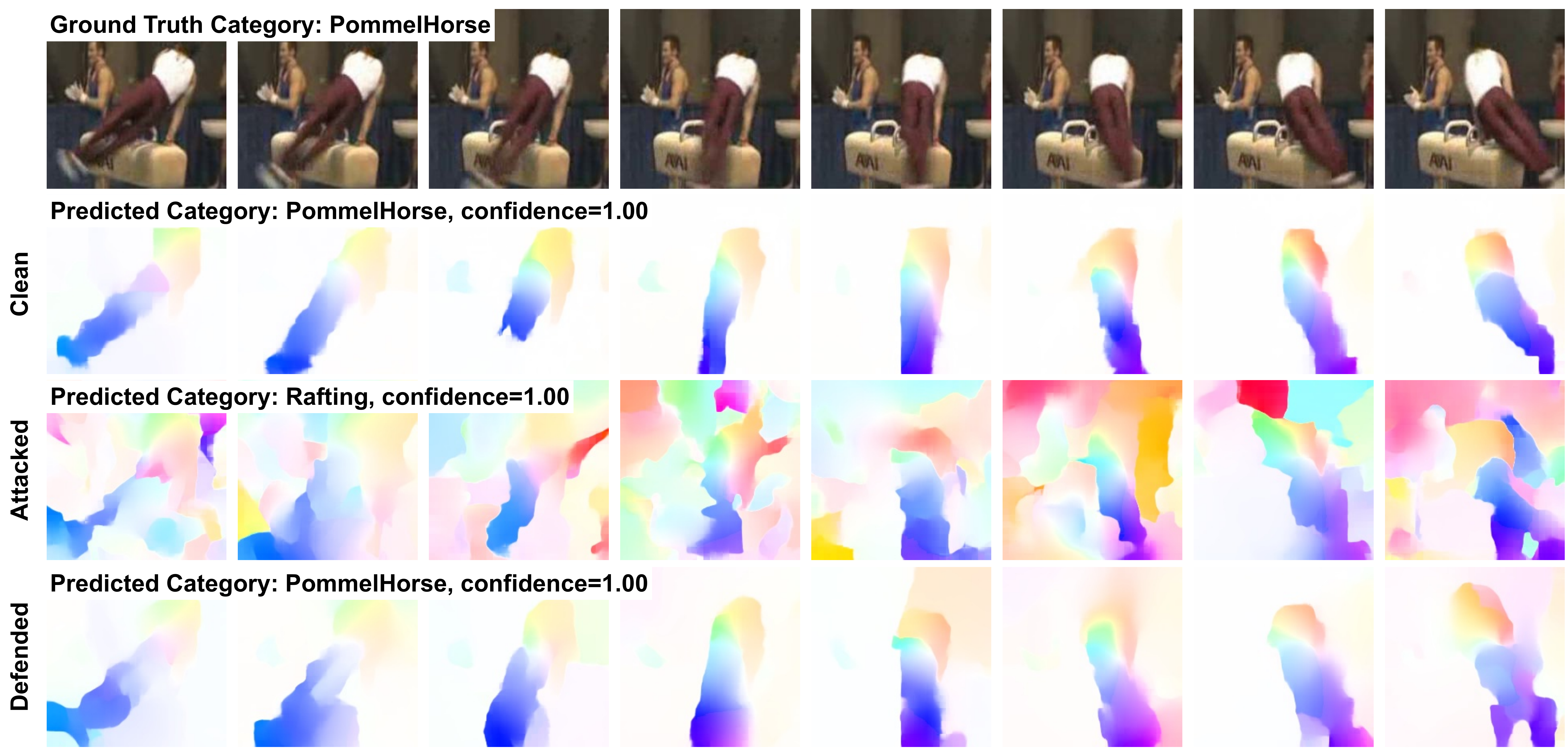}
  \label{fig:flpw}
  \vspace{-7mm}
\end{figure*}

\begin{figure*}[!h]
\centering
   \includegraphics[width=0.9\textwidth]{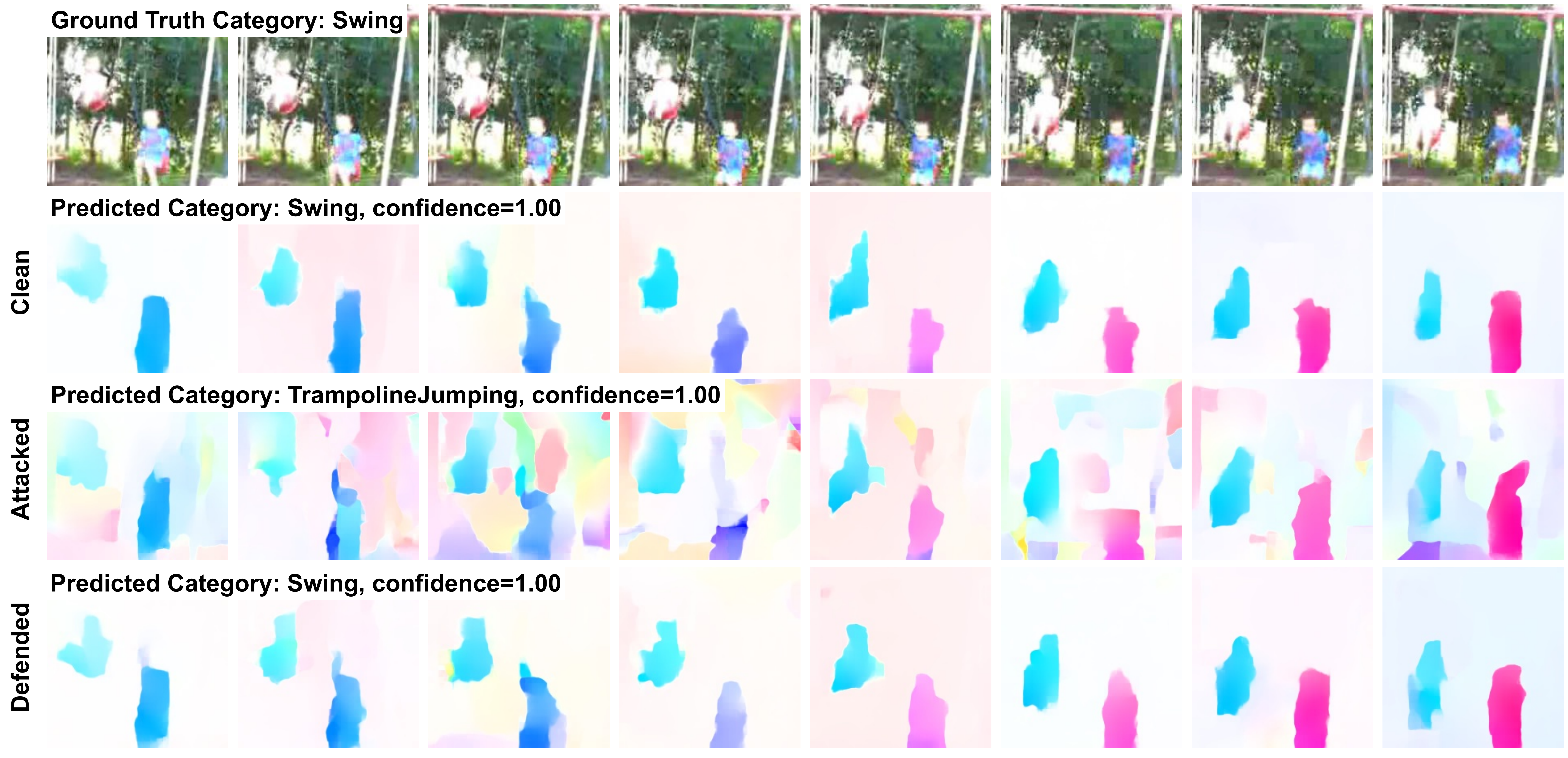}
  \label{fig:flpw}
  \vspace{-7mm}
\end{figure*}

\begin{figure*}[!t]
  \vspace{-10mm}
\centering
   \includegraphics[width=0.9\textwidth]{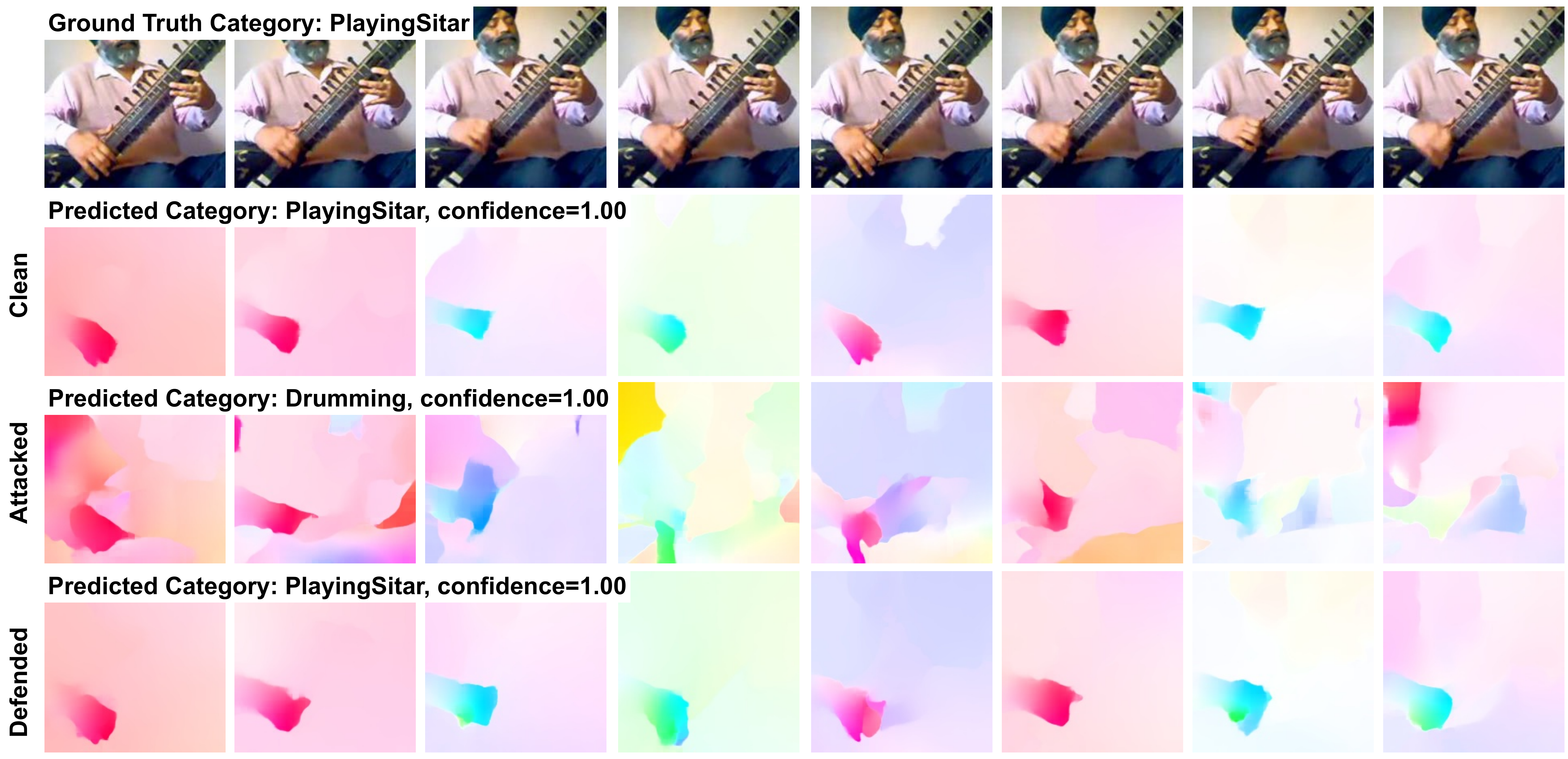}
  \label{fig:flpw}

\end{figure*}

\clearpage



\end{document}